\documentclass[12pt]{article} 

\usepackage[T1]{fontenc}
\usepackage{latexsym}
\usepackage{graphicx}
\usepackage{amsmath}
\usepackage{theorem}
\usepackage{epsfig}
\usepackage{url}

\newtheorem{theorem}{Theorem}[section] 
\newtheorem{corollary}[theorem]{Corollary} 
\newtheorem{definition}[theorem]{Definition} 
\newtheorem{proposition}[theorem]{Proposition} 
\newtheorem{lemma}[theorem]{Lemma} 

\theorembodyfont{\rm}
\newtheorem{example}[theorem]{Example} 
\newtheorem{remark}[theorem]{Remark}

\newcommand{\ebox}{\mbox{}\nobreak\hfill\hspace{6pt}$\Box$}

\newenvironment{proof}[0]{\noindent\textbf{PROOF.}}{\ebox}

\DeclareMathSymbol{\naf}{\mathord}{symbols}{"18}

\newcommand{\union}{\cup}
\newcommand{\Union}{\bigcup}
\newcommand{\isect}{\cap}
\newcommand{\pair}[2]{\langle{#1},{#2}\rangle}
\newcommand{\rg}[3]{#1#2\dots#2#3}
\newcommand{\set}[1]{\{{#1}\}}
\newcommand{\eset}[2]{\{{#1},\ldots,{#2}\}}
\newcommand{\sel}[2]{\{{#1}\,|\,{#2}\}}
\newcommand{\func}[3]{{#1}:{#2}\rightarrow{#3}}

\newcommand{\true}{\mathbf{t}}
\newcommand{\false}{\mathbf{f}}
\newcommand{\undef}{\mathbf{u}}
\newcommand{\True}[1]{{#1}^{\true}}
\newcommand{\False}[1]{{#1}^{\false}}
\newcommand{\Undef}[1]{{#1}^{\undef}}
\newcommand{\Lor}[1]{\bigvee{#1}}
\newcommand{\val}[2]{{#1}({#2})}

\newcommand{\hb}[1]{\mathrm{Hb}(#1)}
\newcommand{\at}[1]{\mathsf{#1}}
\newcommand{\IF}{\leftarrow}
\newcommand{\END}{;\;\;}
\newcommand{\pot}[1]{{#1}^{\bullet}}
\newcommand{\GLred}[2]{{#1}^{#2}}
\newcommand{\red}[2]{{#1}_{#2}}

\newcommand{\UFi}{UF1}
\newcommand{\UFii}{UF2}
\newcommand{\UFiii}{UF3}

\newcommand{\sys}[1]{\textsc{#1}}

\newcommand{\nlp}[1]{#1_\mathrm{N}}
\newcommand{\dlp}[1]{#1_\mathrm{D}}
\newcommand{\heads}[1]{\mathrm{Heads}(#1)}

\newcommand{\tr}[2]{\mathrm{Tr}_{\mathrm{#1}}(#2)}

\newcommand{\genO}[1]{{\mathrm{G0}}(#1)}
\newcommand{\genI}[1]{{\mathrm{G1}}(#1)}
\newcommand{\support}[1]{{\mathrm{Supp}}(#1)}
\newcommand{\supp}[1]{{#1^{\mathrm{s}}}}

\newcommand{\gen}[1]{{\mathrm{Gen}}(#1)}
\newcommand{\test}[2]{{\mathrm{Test}}(#1,#2)}

\newcommand{\expand}{\mathit{expand}}
\newcommand{\extend}{\mathit{extend}}
\newcommand{\smodels}{\mathit{smodels}}
\newcommand{\gnt}{\mathit{gnt}}
\newcommand{\conflict}{\mathit{conflict}}
\newcommand{\heuristic}{\mathit{heuristic}}
\newcommand{\isminimal}{\mathit{minimal}}
\newcommand{\keyw}[1]{\textbf{#1}}
\newcommand{\GP}{G}

\newcommand{\NP}{\mathrm{NP}}

\begin{document} 

\title{\vspace{-1\baselineskip}\
Unfolding Partiality and Disjunctions \\
in Stable Model Semantics\thanks{\
A preliminary version of this paper \cite{JNSY00:kr} appears in
the Proceedings of the 7th International Conference on
the Principles of Knowledge Representation and Reasoning,
KR'2000.}}

\normalsize
\author{
Tomi Janhunen and Ilkka Niemelä \\
Department of Computer Science and Engineering \\
Helsinki University of Technology \\
P.O.Box 5400, FIN-02015 HUT, Finland \\
\{Tomi.Janhunen,Ilkka.Niemela\}@hut.fi \\
\ \\
Dietmar Seipel \\
University of Würzburg \\
Am Hubland, D-97074 Würzburg, Germany \\
seipel@informatik.uni-wuerzburg.de \\
\ \\
Patrik Simons \\
Neotide Oy \\
Wolffintie 36 \\
FIN-65200 Vaasa, Finland \\
Patrik.Simons@neotide.fi \\
\ \\
Jia-Huai You \\
Department of Computing Science \\
University of Alberta \\
Edmonton, Alberta, Canada T6G 2H1\\
you@cs.ualberta.ca
}
\date{}

\maketitle

\begin{abstract}
The paper studies an implementation methodology for partial and
disjunctive stable models where partiality and disjunctions are
unfolded from a logic program so that an implementation of stable
models for normal (disjunction-free) programs can be used as the core
inference engine.  The unfolding is done in two separate steps.
Firstly, it is shown that partial stable models can be captured by
total stable models using a simple linear and modular program
transformation.  Hence, reasoning tasks concerning partial stable models can
be solved using an implementation of total stable models.  Disjunctive
partial stable models have been lacking implementations which now
become available as the translation handles also the disjunctive case.
Secondly, it is shown how total stable models of disjunctive programs
can be determined by computing stable models for normal programs.
Hence, an implementation of stable models of normal programs can be
used as a core engine for implementing disjunctive programs.  The
feasibility of the approach is demonstrated by constructing a system
for computing stable models of disjunctive programs using the
\sys{smodels} system as the core engine.  The performance of the resulting
system is compared to that of \sys{dlv} which is a state-of-the-art
system for disjunctive programs.
\end{abstract}


\section{INTRODUCTION}

Implementation techniques for declarative semantics of logic programs
have advanced considerably during the last years.  For example, the
XSB system~\cite{SSW96:jicslp} is a WAM-based full logic programming system
supporting the well-founded semantics.  In addition to this kind of a
skeptical approach that is based on query evaluation also a credulous
approach focusing on computing models of logic programs is gaining
popularity. This work has been centered around the stable model
semantics~\cite{GL88:iclp,GL91:ngc}.
There are reasonably efficient implementations
available for computing stable models for disjunctive and normal
(disjunction-free) programs, e.g.,
\sys{dlv}~\cite{DLV},
\sys{smodels}~\cite{SMODELS,SNS02:aij},
\sys{cmodels}~\cite{CMODELS}, and
\sys{assat}~\cite{LZ02:aaai}.
The implementations have provided a basis for a new paradigm for logic
programming called \emph{answer set programming} (a term coined by
Vladimir Lifschitz).  
The basic idea is that a problem
is solved by devising a logic program such that the stable models
of the program provide the answers to the problem, i.e.,
solving the problem is reduced to a stable model computation
task~\cite{Lifschitz99:iclp,MT99:lpp,Niemela99:amai,EGM97:acmtds,BLR97:lpnmr}. 
This approach has led to interesting applications in areas such as
planning~\cite{DNK97:planning,EFLPP2000:cl,BBGBW01:padl}, model
checking~\cite{LRS98:tacas,Heljanko99:fi}, and software
configuration~\cite{Syrjanen00:cl}.

This paper addresses two issues in the stable model semantics:
partiality and disjunctions.  The idea is to develop methodology such
that efficient procedures for computing (total) stable models that are
emerging can be exploited when dealing with partial stable models and
disjunctive programs.
Sometimes it is
natural to use partial stable models to represent a domain. Even when
working with total stable models, partial stable models could be
useful, e.g., for debugging purposes to show what is wrong in a
program without any total stable models.  However, little has been
done on implementing the computation of partial stable models and most
of the work has focused on query evaluation w.r.t.\ the well-founded
semantics.  In the paper we show that total stable models can capture
partial stable models using a simple linear program transformation.
This transformation works also in the disjunctive case showing that
implementations of total stable models, e.g.\ \sys{dlv}, can be used for
computing partial stable models. Using a suitable transformation of
queries, a mechanism for query answering can be realized as well.

Our translation is interesting in many respects.  First, it should be
noted that the translation does not follow directly from the
complexity results already available.  It has been shown, e.g., that
the problem of deciding whether a query is contained in some model
(possibility inference) is $\Sigma^p_2$-complete for both partial
and total stable models~\cite{EG95:amai,ELS98:tcs}.  This
implies that there exists a polynomial time reduction from possibility
inference w.r.t.\ partial models to possibility inference w.r.t.\
total models. However, this kind of a translation is guaranteed to
preserve only the yes/no answer to the possibility inference problem.
Second, not all translations are satisfactory from a
computational point of view.  In practice, when a program is compiled
into another form to be executed, certain computational properties of
the translation play an important role:
\begin{itemize}
\item
efficiency of the compilation (in which order of polynomial), 
\item
modularity (are independent, separate  compilations of parts of a
program possible), and 
\item
structural preservation (are the composition and intuition of the
original program  preserved so that debugging and understanding of
runtime behavior are made possible). 
\end{itemize}
All this points to the importance of finding good translation methods
to enable the use of an existing inference engine to solve other
interesting problems.

The efficiency of procedures for computing stable models of normal
programs has increased substantially in recent years. An interesting
possibility to exploit the computational power of such a procedure is
to use it as a {\em core engine} for implementing other reasoning
systems.
In this paper, we follow this approach and develop a method for
reducing stable model computation of disjunctive programs to the
problem of determining stable models for normal programs.
This is non-trivial as deciding whether a disjunctive program has a
stable model is $\Sigma^p_2$-complete~\cite{EG95:amai} whereas the
problem is $\NP$-complete in the non-disjunctive
case~\cite{MT91:jacm}.
The method has been implemented using the \sys{smodels} system
\cite{SMODELS,SNS02:aij} as the core engine.  The performance of the
implementation is compared to that of \sys{dlv}, which is a
state-of-the-art system for computing stable models for disjunctive
programs.

There are a number of novelties in the work.  Maximal partial stable
models for normal programs are known as {\em regular models}, {\em
M-stable models}, and {\em preferred
extensions}~\cite{Dung95:jlp,Sacca90:pods,YY94:jcss}.  Although this semantics
has a sound and complete top-down query answering
procedure~\cite{Dung95:jlp,EK89:iclp,LY01:ijcai}, so far very little effort
has been given to a serious implementation. For disjunctive programs,
to our knowledge, no implementation has ever been attempted.  As a
result, we obtain (perhaps) the first scalable implementation of the
regular model/preferred extension semantics, and the first
implementation ever for partial stable model semantics for disjunctive
programs.
Our technical work on the relationship between stable and partial
stable models via a translational approach provides a compelling
argument for the naturalness of partial stable models: stable models
and partial stable models share the same notion of unfoundedness,
carefully studied earlier in~\cite{ELS97:amai,LRS97:ic}.
Finally, we demonstrate how key tasks in computing 
disjunctive stable models can be reduced to stable model computation for
normal programs by suitable program transformations. 
In particular, we develop techniques for mapping a disjunctive program
into a normal one such that the set of stable models of the normal
program covers the set of stable models of the disjunctive one and 
in many case even coincides with it. 
Moreover, we devise a method where the stability of a 
model candidate for a disjunctive program can be determined by
transforming the disjunctive program into a normal one and checking the
existence of a stable model for it.
Finally, in the experimental part of this paper, we present a new way
of encoding quantified Boolean formulas as disjunctive logic programs.
This transformation is more economical in the number of propositional
atoms and disjunctive rules than earlier transformations presented in
the literature
\cite{EG95:amai,LPFEGPS03:submitted}.

The rest of the paper is structured as follows.
We first review the basic definitions and concepts
in Section~\ref{section:definitions}.
It is then shown in Section~\ref{section:partiality} that partial
stable models can be captured with total stable models using a simple
program transformation.
In Section~\ref{section:disjunctions}, we describe the method for
computing disjunctive stable models using an implementation of
non-disjunctive programs as a core engine.
After this, we present some experimental results in
Section~\ref{section:experiments} and finish with concluding
remarks in Section~\ref{section:conclusions}.

As a comment on the historical development of the translation given in
Section 3, the characterization of partial stable models as stable
models of the transformed program was first sketched for normal
programs in a proof by Schlipf \cite[Theorem 3.2]{Schlipf95:jcss}.
For disjunctive programs, it was discovered and proven in
\cite{SMR97:ilps}, and independently in \cite{JNSY00:kr}.  In the
current paper we present a proof based on unfounded sets, which was
given in \cite{JNSY00:kr}, as this proof reveals some of the
properties of unfounded sets which are of interest in their own right.
Yet another approach to computing the partial stable models of a
disjunctive program based on a program transformation has been
developed by Ruiz and Minker \cite{RM95:nmelp}: a disjunctive program
$P$ is translated into a positive disjunctive program $P^{3S}$ with
constraints, the {\em 3S--transformation} of $P$, such that the total
minimal models of $P^{3S}$ that additionally fulfill the constraints
coincide with the partial stable models of $P$.


\section{DEFINITIONS AND NOTATIONS} 
\label{section:definitions}

A {\em disjunctive logic program} $P$ (or, just {\em disjunctive
program} $P$) is a set of rules of the form
\begin{equation}
\rg{a_1}{\lor}{a_k}\IF\rg{b_1}{,}{b_m},\rg{\naf c_1}{,}{\naf c_n}
\label{eq:drule}
\end{equation}
where $k \geq 1$, $m,n \geq 0$ and $a_i$'s, $b_i$'s and $c_i$'s are
atoms from the Herbrand base $\hb{P}$\footnote{For the sake of
convenience, we assume that a given program $P$ is already
instantiated by the underlying Herbrand universe, and is thus ground.}
of $P$.
Let us also distinguish subclasses of disjunctive programs. If $k=1$
for each rule of $P$, then $P$ is a {\em disjunction-free} or {\em
normal program}. If $n=0$ for each rule of $P$, then $P$ is called
{\em positive}.

{\em Literals} are either atoms from $\hb{P}$ or expressions of the
form $\naf a$ where $a\in\hb{P}$.  For a set of atoms
$A\subseteq\hb{P}$, we define $\naf A$ as $\sel{\naf a}{a\in A}$.  Let
us introduce a shorthand $A\IF B,\naf C$ for rules where
$A\neq\emptyset$, $B$ and $C$ are subsets of $\hb{P}$.  In harmony
with (\ref{eq:drule}), the set of atoms $A$ in the {\em head} of the
rule is interpreted disjunctively while the set of literals
$B\union\naf C$ in the {\em body} of the rule is interpreted
conjunctively.
We wish to further simplify the notation $A\IF B,\naf C$ in some
particular cases. When $A$, $B$ or $C$ is a singleton $\set{a}$, we
write $a$ instead of $\set{a}$. If $B=\emptyset$ or $C=\emptyset$ we
omit $B$ and $\naf C$ (respectively) as well as the separating comma
in the body of the rule.


\subsection{PARTIAL AND TOTAL MODELS}

We review the basic model-theoretic concepts by following the
presentation in \cite{ELS98:tcs}.
Let $P$ be any disjunctive program. A {\em partial interpretation} $I$
for $P$ is a pair $\pair{T}{F}$ of subsets of $\hb{P}$ such that
$T\isect F=\emptyset$.  The atoms in the sets $\True{I}=T$,
$\False{I}=F$ and $\Undef{I}=\hb{P}-(T\union F)$ are considered to be
{\em true}, {\em false}, and {\em undefined}, respectively. We
introduce constants $\true$, $\false$, and $\undef$, to denote the
respective three truth values.
A partial interpretation $I$ for $P$ is a {\em total} interpretation
for $P$ whenever $\Undef{I}=\emptyset$, i.e., if every atom of
$\hb{P}$ is either true or false. When no confusion arises, we use
$\True{I}$ alone to specify a total interpretation $I$ for $P$ (then
$\False{I}=\hb{P}-\True{I}$ and $\Undef{I}=\emptyset$ hold).

Given a partial interpretation for $P$, the truth values of atoms are
determined by $\True{I}$, $\False{I}$ and $\Undef{I}$ as explained
above while $\true$, $\false$ and $\undef$ have their fixed truth
values. For more complex logical expressions $E$, we use $\val{I}{E}$
to denote the truth value of $E$ in $I$. The value $\val{I}{\naf a}$
is defined to be $\true$, $\false$, or $\undef$ whenever $\val{I}{a}$
is $\false$, $\true$, or $\undef$, respectively.
To handle conjunctions and disjunctions, we introduce an ordering on
the three truth values by setting $\false<\undef<\true$. By default, a
set of literals $L=\eset{l_1}{l_n}$ denotes the conjunction
$\rg{l_1}{\land}{l_n}$ while $\Lor{L}$ denotes the corresponding
disjunction $\rg{l_1}{\lor}{l_n}$. The truth values $\val{I}{L}$ and
$\val{I}{\Lor{L}}$ are defined as the respective minimum and maximum
among the truth values $\rg{\val{I}{l_1}}{,}{\val{I}{l_n}}$.
A rule $A\IF B,\naf C$ is satisfied in $I$ if and only if
$\val{I}{\Lor{A}}\geq\val{I}{B\union\naf C}$.
A partial interpretation $M$ for $P$ is a {\em partial model} of $P$
if all rules of $P$ are satisfied in $M$, and for a {\em total model},
also $\Undef{M}=\emptyset$ holds. Let us then introduce an ordering
among partial models of a disjunctive program: $M_1\leq M_2$ if and only if
$\True{M_1}\subseteq\True{M_2}$ and $\False{M_1}\supseteq\False{M_2}$.
A partial model $M$ of $P$ is a {\em minimal} one if there is no
partial model $M'$ of $P$ such that $M'<M$ (i.e., $M'\leq M$ and
$M'\neq M$). In case of total models, we have $N_1\leq N_2$ if and only if
$N_1\subseteq N_2$. Moreover, a total model $N$ of $P$ is considered
to be a {\em minimal} one if there is no total model $N'$ of $P$ such
that $N'\subset N$.


\subsection{STABLE MODELS}

Given a partial interpretation $I$ for a disjunctive program $P$, we
define a reduction of $P$ as follows:
\[\GLred{P}{I}=
 \sel{A\IF B}
     {A\IF B,\naf C\in P\mbox{ and }
      C\subseteq\False{I}} .
\]
Note that this transformation coincides with the {\em
Gelfond-Lifschitz reduction} of $P$ (the GL-reduction of $P$) when $I$
is a total interpretation.

\begin{definition}[Total stable model]
\label{def:stablemodels}
A total interpretation $N$ for a disjunctive program $P$ is a stable
model if and only if $N$ is a minimal total model of $\GLred{P}{N}$.
\end{definition}

The original definition of partial stable models \cite{Przymusinski90:iclp,Przymusinski90:fi}
is based on a weaker reduction. Given a disjunctive program $P$ and an
interpretation $I$, the reduction $\red{P}{I}$ is the set of rules
obtained from $P$ by replacing any $\naf c$ in the body of a rule by
$I(\naf c)$.
As noted in \cite{Przymusinski90:iclp}, the only practical difference between
$\GLred{P}{I}$ and $\red{P}{I}$ is that $\red{P}{I}$ has rules that
correspond to rules of $A\IF B,\naf C\in P$ satisfying
$\val{I}{\naf C}=\undef$. Note that if $\val{I}{\naf C}=\true$, then
$A\IF B\in\GLred{P}{I}$, and if $\val{I}{\naf C}=\false$, then
the partial models of $\red{P}{I}$ are not constrained by the rule
included in $\red{P}{I}$.

\begin{definition}[Partial stable model]
\label{def:partial-stable}
A partial interpretation $M$ for a disjunctive program $P$ is a
partial stable model of $P$ if and only if $M$ is a minimal partial model of
$\red{P}{M}$.
\end{definition}

In the above definition, the relation between $M$ and $\red{P}{M}$ is
similar to the one for the total stable model, both for the purpose of
preserving the stability condition. While maximizing falsity and
minimizing true atoms, a partial stable model does not insist that
every atom must be either true or false.
(Partial) stable models are intimately related to {\em unfounded
sets}~\cite{ELS97:amai,LRS97:ic}. 

\begin{definition}[Unfounded sets]
\label{def:unfounded-sets}
Let $I$ be a partial interpretation for a disjunctive program $P$.
A set $U\subseteq\hb{P}$ of ground atoms is an {\em unfounded set} for
$P$ w.r.t.\  $I$, if at least one of the following conditions holds for
each rule $A\IF B,\naf C\in P$ such that $A\isect U\neq\emptyset$:
\begin{description}
\item[\textnormal{\UFi:}]
$B\isect\False{I}\neq\emptyset$ or $C\isect\True{I}\neq\emptyset$,
\item[\textnormal{\UFii:}]
$B\isect U\not=\emptyset$, or
\item[\textnormal{\UFiii:}]
$(A - U)\isect(\True{I}\union\Undef{I})\neq\emptyset$.
\end{description}
An unfounded set $U$ for $P$ w.r.t.\ $I$ is $I$-consistent if
and only if $U\isect\True{I}=\emptyset$.
\end{definition}
The conditions \UFi\ and \UFiii\ above coincide with the conditions
\begin{center}
$\val{I}{B\union\naf C}=\false$ and $\val{I}{\Lor(A-U)}\neq\false$,
\end{center}
respectively.
The intuition is that the atoms of an unfounded set $U$ can be assumed
to be false without violating the satisfiability of any rule $A\IF
B,\naf C$ of the program whose head contains some atoms of $U$.  For
any such rule, either
the rule body is false in $I$ (\UFi),
or the rule body can be falsified by falsifying the atoms in $U$ (\UFii),
or the head of the rule is not false in $I$ (\UFiii).
In particular, unfounded sets w.r.t.\ partial/total models can be used
for constructing smaller partial/total models (recall the definition
of minimal partial and total models) in a way that is made precise
by what follows.

\begin{lemma}
\label{lemma:set-unfounded-false}
Let $M=\pair{T}{F}$ be a partial model of a positive disjunctive
program $P$ and $U$ an unfounded set for $P$ w.r.t.\ $M$.
Then, if $M$ is total or $U$ is $M$-consistent,
$M'=\pair{T-U}{F\union U}$ is a partial model of $P$.
\end{lemma}

\begin{proof}
Let $M$, $P$, $U$ and $M'$ be defined as above. Additionally,
we assume that
(a)
$M$ is total, or
(b)
$U$ is $M$-consistent.
Let us then assume that some rule $A\IF B$ of $P$ is not
satisfied in $M'$ which means that
$\val{M'}{\Lor A}<\val{M'}{B}$.
Thus
(i)
$\val{M'}{\Lor A}<\true$ and $\val{M'}{B}=\true$, or
(ii)
$\val{M'}{\Lor A}=\false$ and $\val{M'}{B}=\undef$.
Our proof splits in two separate threads.

\begin{itemize}

\item[{\bf I.}]
Assume that $A\isect U=\emptyset$ holds.
Consider the case (i).
Now $\val{M'}{\Lor A}<\true$ implies $A\isect(T-U)=\emptyset$.  Since
$A\isect U=\emptyset$ holds, too, we obtain $A\isect T=\emptyset$ so
that $\val{M}{\Lor A}<\true$. On the other hand, $\val{M'}{B}=\true$
implies $B\subseteq T-U$. Thus $B\subseteq T$ and $\val{M}{B}=\true$
holds as well. But then $\val{M}{\Lor A}<\val{M}{B}$, a contradiction.
The case (ii) is analyzed next.
Now $\val{M'}{\Lor A}=\false$ implies $A\subseteq F\union U$ as well
as $A\subseteq F$, since $A\isect U=\emptyset$.  Thus $\val{M}{\Lor
A}=\false$. Moreover, from $\val{M'}{B}=\undef$ we obtain
$B\isect(F\union U)=\emptyset$. Thus we obtain $B\isect
F\neq\emptyset$ so that $\val{M}{B}>\false$ holds.
To conclude, we have established that
$\val{M}{\Lor A}<\val{M}{B}$, a contradiction.

\item[{\bf II.}]
Otherwise $A\isect U\neq\emptyset$ holds. Then at least one of the
unfoundedness conditions is applicable to $A\IF B$. If \UFi\ is,
$\val{M}{B}=\false$ holds. It follows that $B\subseteq F$ and
$B\subseteq F\union U$. Thus $\val{M'}{B}=\false$ contradicting both
(i) and (ii).
If \UFii\ is applicable, we have $B\isect U\neq\emptyset$. It follows
that $B\isect(F\union U)\neq\emptyset$ so that $\val{M'}{B}=\false$, a
contradiction.

Thus \UFiii\ must apply, i.e., $\val{M}{\Lor(A-U)}>\false$ holds.
Let us then consider cases (a) and (b) separately.
\begin{enumerate}
\item[(a)]
If $M$ is total, we have necessarily $\val{M}{\Lor(A-U)}=\true$. This
implies that some atom $a\in A-U$ belongs to $T$. Thus also $a\in T-U$
and $\val{M'}{\Lor A}=\true$, a contradiction with both (i) and (ii).

\item[(b)]
If $U$ is $M$-consistent, we have $U\isect T=\emptyset$. By
$\val{M}{\Lor(A-U)}=\true$ there is an atom $a\in A-U$ such that
$a\not\in F$.  Then $a\not\in F\union U$ which implies $\val{M'}{\Lor
A}>\false$. Thus (ii) is impossible and (i) implies $\val{M'}{\Lor
A}=\undef$ and $\val{M'}{B}=\true$. It follows that
$A\isect(T-U)=\emptyset$ and $B\subseteq T-U$. Since $U\isect
T=\emptyset$, the former implies $A\isect T=\emptyset$ while the
latter implies that $B\subseteq T$. Consequently, $\val{M}{\Lor
A}<\true$ and $\val{M}{B}=\true$, i.e., $\val{M}{\Lor A}<\val{M}{B}$,
a contradiction.
\end{enumerate}
\end{itemize}
\end{proof}

Let us yet emphasize the content of Lemma
\ref{lemma:set-unfounded-false} when $M$ is total (and $U$ need not be
$M$-consistent).  Then $M'=M-U$ is also a total model of $P$.
A couple of examples on unfounded sets follow.

\begin{example}
Consider a disjunctive program 
\[
P =\{ a \lor b \IF c, \naf a\}
\]
and an interpretation $I =\langle \emptyset, \{a\}\rangle$.  The only
rule in $P$ has its body undefined in $I$, hence \UFi\ is not
applicable. The set $\{a\}$ is unfounded w.r.t.\ $I$ since $b$ is
undefined in $I$ and not in the set, hence \UFiii\ is applicable.
On the other hand, the set $\{b\}$ is not unfounded w.r.t.\ $I$ whereas
$\{c\}$ is unfounded w.r.t.\ $I$.
Once $c$ belongs to an unfounded set, the atoms $a$ and $b$ can both
get in due to \UFii. Hence, we have $U =\{a,b,c\}$ as an unfounded set
w.r.t.\ $I$.

Comparing $U$ with $\False{I}$, we find that $\False{I}$ does not
maximize the atoms that should be false.  This program $P$ has exactly
one stable model (which is also a partial stable model) in which all
three atoms are false.
\ebox
\end{example}

Unlike the case for normal programs, the union of unfounded sets may
not be an unfounded set.
\begin{example}
Consider a program $P$ containing only one rule
\[
a \lor b \IF
\]
and an interpretation $I =\langle \{a,b\},\emptyset\rangle$.  The
program has two non-empty unfounded sets w.r.t.\ $I$, $\{a\}$ and
$\{b\}$.  Either $a$ or $b$ depends on the other one not in the set
for \UFiii\ to be applicable. However, \UFiii\ becomes not applicable when
both $a$ and $b$ are in, thus the union $\{a,b\}$ is not an unfounded
set.
\ebox
\end{example}

An interpretation $I$ for $P$ becomes particularly interesting when
the union of all unfounded sets $U$ for $P$ w.r.t.\ $I$ is also an
unfounded set for $P$ w.r.t.\ $I$. In this case, the program $P$
possesses {\em the greatest unfounded} set $U$ for $P$ w.r.t.\ $I$.

\begin{definition}
\label{def:unfounded-free}
A total interpretation $I$ is said to be {\em unfounded free} for a
program $P$ if and only if there is no unfounded set $U$ for $P$
w.r.t.\ $I$ such that $U\isect\True{I}\neq\emptyset$.
\end{definition}

The notion of unfounded freeness captures the stable model
beautifully.
\begin{theorem} {\bf \cite{LRS97:ic}}
\label{theorem:SM-character}
Let $M$ be a total interpretation for a disjunctive program $P$.
Then, the following are equivalent
\begin{itemize} 
\item
$M$ is a stable model of $P$.
\item 
$\False{M}$ is the greatest unfounded
set for $P$ w.r.t.\ $M$.
\item
$M$ is unfounded free for $P$.
\end{itemize} 
\end{theorem} 

On the other hand, Eiter et al. \cite{ELS97:amai} show that partial
stable models can be defined essentially without reference to
three-valued logic.

\begin{theorem} {\bf \cite{ELS97:amai}}
\label{theorem:PSM-character}
If $M$ is a partial interpretation for a disjunctive program $P$,
then $M$ is a partial stable model of $P$ if and only if

\begin{itemize}
\item
$\True{M}$ is a minimal total model of $\GLred{P}{M}$ and
\item
$\False{M}$ is a maximal $M$-consistent unfounded set for $P$ w.r.t.\ $M$.
\end{itemize}
\end{theorem}

The first condition in the theorem is called {\em foundedness}
in \cite{ELS97:amai}.
The differences between these two theorems are quite subtle.  The
strictness of stable models enforces a simpler relationship between
stable models and unfounded sets. Therefore, neither maximality nor
consistency nor foundedness need be explicitly stated. The characterization
of partial stable models in Theorem~\ref{theorem:PSM-character}
accounts for a more reflexible situation: since $M$ may not be a total
model, maximality should extend the set of false atoms as much as
possible without causing inconsistency.  However, maximality and
consistency are still not strong enough.
\begin{example}
Consider a disjunctive program
\[
P = \{ a \lor b \IF \naf a\}
\]
and an interpretation $I = \langle \{a\}, \{b\} \rangle$ which is
total so that the definition of unfoundedness makes no difference in
valuation under $I$.  Since the body of the rule is false in $I$, $U_1
= \{a\}$, $U_2 = \{b\}$, and $U_3 = \{a,b\}$ are all nonempty
unfounded sets in this case. It follows immediately by Theorem
\ref{theorem:PSM-character} that $I$ is not a stable model. However,
$U_2$ is maximally $I$-consistent yet it is not a partial stable model
because $\True{I}$ is not a minimal model of $P^I$.

Note that $P$ has a unique (partial) stable model, which is $\langle
\{b\},\{a\}\rangle$.
\ebox
\end{example}

The characterizations for partial and total stable models in terms of
unfounded sets provide a powerful tool for establishing relationships
between stable models and partial stable models.


\section{UNFOLDING PARTIALITY}
\label{section:partiality}

In this section, we first show a translation for a disjunctive program
into another disjunctive program.  We then prove that the translation
preserves the semantics of partial stable models. This result allows
us to compute the partial stable models of a program by computing the
stable models of the translated program.  Finally we address the
problem of query answering under the translation.

\subsection{TRANSLATION}

Let $P$ be a disjunctive program. In the following, we describe a
translation of $P$ into another disjunctive program $\tr{}{P}$ such
that the stable models of $\tr{}{P}$ correspond to the partial stable
models of $P$.

Let us introduce a new atom $\pot{a}$ for each $a\in\hb{P}$.  An atom
$\pot{a}$ is said to be {\em marked}, and an ordinary atom $a$ is then
said to be {\em unmarked}. The intuitive reading of $\pot{a}$ is that
$a$ is {\em potentially} true.
For a set of literals $L \subseteq\hb{P} \cup \naf\hb{P}$, we define
$\pot{L}=\sel{\pot{a}}{a\in L} \union \sel{\naf \pot{a}}{\naf a\in L}$.
The translation $\tr{}{P}$ of a disjunctive program $P$ is as
follows:
\begin{multline}
\tr{}{P}=
\sel{A\IF B,\naf\pot{C}\END\pot{A}\IF\pot{B},\naf C}{A\IF B,\naf C\in P}
\ \union \\
\sel{\pot{a}\IF a}{a\in\hb{P}}
\end{multline}
where semicolons are used to separate program rules.
Note that the Herbrand base of $\hb{\tr{}{P}}$ is
$\hb{P}\union\pot{\hb{P}}$. The rules $\pot{a}\IF a$ introduced
for each $a\in\hb{P}$ enforce consistency in the sense that if $a$ is
true, then $a$ must also be potentially true.

\begin{remark}
Although for presentational purposes the translation is defined for
ground programs, exactly the same translation applies to non-ground
programs as well: for each predicate $p$ we introduce a new predicate
$\pot{p}$, hence for a (ground or non-ground) atom $\phi =
p(\rg{t_1}{,}{t_n})$, the new atom is $\pot{\phi} =
\pot{p}(\rg{t_1}{,}{t_n})$ (cf. Example~\ref{barber}).  Since our
proofs do not depend on the assumption that a given program is finite,
the conclusions reached cover also any non-ground program with
function symbols whose semantics is determined by treating the program
as a shorthand for its (possibly infinite) Herbrand instantation.
\end{remark}

A partial stable model of a given program will be
interpreted by a corresponding stable model of the transformed
program.
The extra symbol $\pot{a}$ for each atom $a$ provides an opportunity
to represent {undefined} (in three-valued logic) in terms of truth
values of $\pot{a}$ and $a$ in two-valued logic. For each pair $a$ and
$\pot{a}$, either of which can be true or false, there are four
possibilities: when $\pot{a}$ and $a$ are in agreement, that is when
they are both true or both false, the truth value of $a$ is their
commonly agreed truth value; the combination where $a$ is false and
$\pot{a}$ is true then represents that $a$ is undefined; and the
fourth possibility where $a$ is true and $\pot{a}$ is false is ruled out by any
models due to the consistency rules.
This intended representation of a partial stable model is given by the
following equations.

\begin{definition} 
\label{def:CE1234}
Let $M$ be a partial interpretation of a program $P$
and $N$ a total interpretation of $\tr{}{P}$.
The interpretations $M$ and $N$ are said to satisfy the {\em
correspondence equations} if and only if the following equations hold.
\begin{align*}
\True{M}  & =
\sel{a\in\hb{P}}{a\in\True{N}\mbox{ and }\pot{a}\in\True{N}}
\tag{CE1} \\ 
\False{M} & =
\sel{a\in\hb{P}}{a\in\False{N}\mbox{ and }\pot{a}\in\False{N}}
\tag{CE2} \\
\Undef{M} & =
\sel{a\in\hb{P}}{a\in\False{N}\mbox{ and }\pot{a}\in\True{N}}
\tag{CE3} \\
\emptyset & =
\sel{a\in\hb{P}}{a\in\True{N}\mbox{ and }\pot{a}\in\False{N}}
\tag{CE4}
\end{align*}
\end{definition}

Note that total interpretations that are models of $\tr{}{P}$ satisfy
CE4 immediately, since the set of rules $\sel{\pot{a}\IF
a}{a\in\hb{P}}$ is included in $\tr{}{P}$. Consequently, the ``fourth
truth value'' is ruled out.
The following example demonstrates how the representation given in
Definition \ref{def:CE1234} allows us to capture the partial stable
models of a disjunctive program $P$ with the total stable models of
$\tr{}{P}$.

\begin{example}
\label{ex:stable-models}
Consider a disjunctive program
\begin{center}
$P=\{a\lor b \IF \naf c\END
     b \IF \naf b\END
     c \IF \naf c\}$.
\end{center}
Now $a$ becomes false by the minimization of partial models, since the
falsity of $a$ does not affect the satisfiability of any rule. Thus
the unique partial stable model of $P$ is
$M=\pair{\emptyset}{\set{a}}$. Note that the reduction
$\red{P}{M}=
 \set{a\lor b\IF\undef\END
      b\IF\undef\END
      c\IF\undef}$.
Then consider the translation
\[
\begin{array}{r@{\,}l@{\,}lll@{}l}
\tr{}{P}= & \{ & a \lor b\IF\naf\pot{c}\END
               & b\IF\naf\pot{b}\END
               & c\IF\naf\pot{c}\END \\
          &    & \pot{a}\lor\pot{b}\IF\naf c\END
               & \pot{b}\IF\naf b\END
               & \pot{c}\IF\naf c\END \\
          &    & \pot{a}\IF a\END
               & \pot{b}\IF b\END
               & \pot{c}\IF c & \}.
\end{array}
\]
The unique stable model of $\tr{}{P}$ is $N=\set{\pot{b},\pot{c}}$
which represents (by CE2 and CE3) the setting that $b$ and $c$ are
undefined and $a$ is false in $M$.
\ebox
\end{example}

It is well-known that a disjunctive program $P$ may not have any
partial stable models. In such cases, the translation $\tr{}{P}$
should not have stable models either, if the translation $\tr{}{P}$ is
to be faithful.
\begin{example}
\label{ex:no-stable-models}
Consider a disjunctive program
\begin{center}
$P=\{a\lor b\lor c\IF\END
     a\IF\naf b\END
     b\IF\naf c\END
     c\IF\naf a\}$
\end{center}
and its translation 
\[
\begin{array}[t]{l@{\,}c@{\,}l@{\,}llll@{\,}ll}
\tr{}{P} & = &
\{ & a\lor b\lor c\IF\END & a\IF\naf \pot{b}\END
                          & b\IF\naf \pot{c}\END & c\IF\naf \pot{a}\END \\
&& & \pot{a}\lor\pot{b}\lor\pot{c}\IF\END & \pot{a}\IF\naf b\END
                          & \pot{b}\IF\naf c\END & \pot{c}\IF\naf a
   & \}\union C
\end{array}
\]
where
$C=\{\pot{a}\IF a\END
     \pot{b}\IF b\END
     \pot{c}\IF c\}$
is the set of consistency rules.

Consider a partial model $M=\pair{\set{a,b}}{\emptyset}$ of $P$ and a
total model $N=\set{a,\pot{a},b,\pot{b},\pot{c}}$ of $\tr{}{P}$ that
satisfy the equations CE1--CE4 in Definition \ref{def:CE1234}.
Now the reduced program $\red{P}{M}$ is 
\[
\{ a \lor b \lor c\IF\END
   a \IF \false\END
   b \IF \undef\END
   c \IF \false \}
\] and since
$M'=\pair{\set{a,b}}{\set{c}}<M$ is a partial model of
$\red{P}{M}$, $M$ is not a partial stable model of $P$.
On the other hand, the reduct 
\[
\GLred{\tr{}{P}}{N}=
\{ a\lor b\lor c\IF\END
   \pot{a}\lor\pot{b}\lor\pot{c}\IF\END
   \pot{b}\IF \}\union C .
\] 
But $N'=\set{a,\pot{a},b,\pot{b}}\subset N$ is a model of
$\GLred{\tr{}{P}}{N}$ so $N$ is not a stable model of $\tr{}{P}$.
The reader may analyze the other candidates in a similar fashion. It
turns out that $P$ does not have partial stable models. Nor does
$\tr{}{P}$ have stable models.
\ebox
\end{example}

Partial stable models can be viewed as a logic programming account of
the solution of semantic paradoxes due to Kripke \cite{Kripke75:jp}.  In
this account, undefined means {\em unknown} for some individuals which
will not lose semantic interpretations for other individuals.
\begin{example}
\label{barber}
Consider the following program with variables:
\[
\begin{array}{l@{\ }c@{\ }lll}
P & = & \{ &
    \at{shave}(\at{bob}, x)
       \IF \naf \at{shave}(x,x)\END\\
&&& \at{pay\_by\_cash}(y,x) \lor \at{pay\_by\_credit}(y,x)
       \IF \at{shave}(x,y)\END\\
&&& \at{accepted}(x,y)
       \IF \at{pay\_by\_cash}(x,y)\END\\
&&& \at{accepted}(x,y)
       \IF \at{pay\_by\_credit}(x,y)\ \}. \\
\end{array}
\]
This program intuitively says that Bob shaves those who do not shave
themselves; if $x$ shaves $y$ then $y$ pays $x$ by cash or by credit;
either way is accepted. The predicate $\at{accepted}$ is used here to
demonstrate disjunctive reasoning.

Assume there is another person, called Greg.  Then clearly, we
should conclude Bob shaves Greg, and Greg pays Bob by cash or
by credit, either way is accepted. However, the program has no stable
models in this case due to the paradox whether Bob shaves
himself or not. But it has two partial stable models, in both of which
$\at{shave}(\at{greg},\at{greg})$ is false and
$\at{shave}(\at{bob},\at{bob})$ is undefined (unknown).
By translating the first two rules of $P$ we obtain
\[
\begin{array}{ll}
\at{shaves}(\at{bob}, x)
   \IF \naf \pot{\at{shaves}} (x,x)\END \\
\pot{\at{shaves}}(\at{bob}, x)
   \IF \naf \at{shaves}(x,x)\END \\
\at{pay\_by\_cash}(y,x) \;\lor\; \at{pay\_by\_credit}(y,x)
   \IF \at{shaves}(x,y)\END \mbox{and} \\
\pot{\at{pay\_by\_cash}}(y,x) \lor \pot{\at{pay\_by\_credit}}(y,x)
   \IF \pot{\at{shaves}}(x,y). \\
\end{array}
\]
The full translation $\tr{}{P}$ yields a Herbrand instantiation over
the universe $\set{\at{bob},\at{greg}}$ which has four total stable
models.  One of them is
\[
\begin{array}{r@{\ }c@{\ }l@{\ }l}
N & = & \{ &
\pot{\at{shaves}}(\at{bob},\at{bob}), \\
&&& \at{shaves}(\at{bob},\at{greg}),\ \pot{\at{shaves}}(\at{bob},\at{greg}), \\
&&& \at{pay\_by\_cash}(\at{greg},\at{bob}),
    \ \pot{\at{pay\_by\_cash}}(\at{greg},\at{bob}), \\
&&& \pot{\at{pay\_by\_credit}}(\at{bob},\at{bob}),
    \ \pot{\at{accepted}}(\at{bob},\at{bob}), \\
&&& \at{accepted}(\at{greg},\at{bob}),\ \pot{\at{accepted}}(\at{greg},\at{bob})
    \ \}.
\end{array}
\]
Hence the fact that $\at{shaves}(\at{bob},\at{bob})$ is undefined in the
corresponding partial stable model $M$ (recall the equations in
Definition \ref{def:CE1234}) is represented by
$\pot{\at{shaves}}(\at{bob},\at{bob})$ being true and
$\at{shaves}(\at{bob},\at{bob})$ being false in $N$.
\ebox
\end{example}


\subsection{CORRECTNESS OF THE TRANSLATION}

The goal of this section is to establish a one-to-one correspondence
between the partial stable models of a disjunctive program $P$ and the
(total) stable models of the translation $\tr{}{P}$.
It is first shown that the correspondence equations CE1--CE4 given in
Definition~\ref{def:CE1234} provide a syntactic way to transform a
partial stable model $M$ of $P$ into a total stable model $N$ of
$\tr{}{P}$ and back. More formally, we have the following theorem in
mind.

\begin{theorem}
\label{theorem:PSM-iff-SM}
Let $M$ be a partial interpretation of a disjunctive program $P$ and
$N$ a total interpretation of the translation $\tr{}{P}$ such that
CE1--CE4 are satisfied.
Then $M$ is a partial stable model of $P$ if and only if $N$ is a
(total) stable model of $\tr{}{P}$.
\end{theorem}

Our strategy to prove Theorem \ref{theorem:PSM-iff-SM} is as follows:
first, in two separate lemmas, we show the correspondence, in each
direction, between unfounded sets for $P$ and $\tr{}{P}$ under $M$ and
$N$, respectively.  These two lemmas are interesting in their own
right as they show very tight conditions under which the two
previously studied notions of unfoundedness
\cite{ELS97:amai,LRS97:ic} are related.  These results will then
be used in the proof of the theorem.
We first state two relatively simple facts.  The first says that that
the GL-transform has no effect on unfoundedness, and the second states
that the translation preserves models via CE1--CE4 in
Definition \ref{def:CE1234}.

\begin{proposition} 
\label{no-effect}
Let $P$ be a disjunctive program and $N$ a total
interpretation for $P$.
Then, $X\subseteq\hb{P}$ is an unfounded set for $P$ w.r.t.\ $N$
if and only if
$X$ is an unfounded set for $\GLred{P}{N}$ w.r.t.\ $N$.
\end{proposition} 

\begin{proof}
Note that $A\IF B\in\GLred{P}{N}$ if and only if there is a rule $A\IF
B,\naf C\in P$ such that $C\subseteq\False{N}$, i.e.,
$C\isect\True{N}=\emptyset$.
Then it holds for any $X\subseteq\hb{P}$ that
\begin{center}
\begin{tabular}{cl}
&  $X$ is not an unfounded set for $P$ w.r.t.\ $N$ \\
$\iff$
&
$\exists A\IF B,\naf C\in P$ such that (1) $A\isect X\neq\emptyset$,
(2) $B\isect\False{N}=\emptyset$, \\ & (3) $C\isect\True{N}=\emptyset$,
(4) $B\isect X=\emptyset$, and (5) $(A-X)\isect\True{N}=\emptyset$ \\
$\iff$
&
$\exists A\IF B\in\GLred{P}{N}$ such that (6) $A\isect X\neq\emptyset$,
(7) $B\isect\False{N}=\emptyset$, \\ & (8) $B\isect
X=\emptyset$, and (9) $(A-X)\isect\True{N}=\emptyset$ \\
$\iff$
&
$X$ is not an unfounded set for $\GLred{P}{N}$ w.r.t.\ $N$.
\end{tabular}
\end{center}
\end{proof}

\begin{proposition}
\label{model-for-M-N}
Let $M$ be a partial interpretation for a disjunctive program $P$ and
$N$ a total interpretation for the translation $\tr{}{P}$.  Assume
$M$ and $N$ satisfy the CEs. Then, $M$ is a partial model of $P$ if
and only if $N$ is a total model of $\tr{}{P}$.
\end{proposition}

\begin{proof}
It follows by the correspondence equations CE1--CE4 in Definition
\ref{def:CE1234} that $M$ is not a partial model of $P$ if and only if
\begin{center}
\begin{tabular}{@{}cl@{}}
&
$\exists$ $A\IF B,\naf C\in P$:\ 
$\val{M}{\Lor{A}}<\val{M}{B\union\naf C}$ \\
$\iff$ &
$\exists$ $A\IF B,\naf C\in P$:\ 
$\val{M}{\Lor{A}}<\true$ and $\val{M}{B\union\naf C}=\true$, {\em or} \\
&
$\exists$ $A\IF B,\naf C\in P$:\ 
$\val{M}{\Lor{A}}=\false$ and $\val{M}{B\union\naf C}=\undef$ \\
$\iff$ &
$\exists$ $A\IF B,\naf\pot{C}\in\tr{}{P}$:\ 
$\val{N}{\Lor{A}}=\false$ and $\val{N}{B\union\naf\pot{C}}=\true$, {\em or} \\
& 
$\exists$ $\pot{A}\IF\pot{B},\naf C\in\tr{}{P}$:\ 
$\val{N}{\Lor{\pot{A}}}=\false$ and $\val{N}{\pot{B}\union\naf C}=\true$ \\
$\iff$ &
$\exists$ $A\IF B,\naf\pot{C}\in\tr{}{P}$:\ 
$\val{N}{\Lor{A}}<\val{N}{B\union\naf\pot{C}}$, {\em or} \\
&
$\exists$ $\pot{A}\IF\pot{B},\naf C\in\tr{}{P}$:\ 
$\val{N}{\Lor{\pot{A}}}<\val{N}{\pot{B}\union\naf C}$
\end{tabular}
\end{center}
which is equivalent to stating that $N$ is not a total
model of $\tr{}{P}$, since the consistency rules in $\tr{}{P}$
are automatically satisfied by CE4.
\end{proof}

Still assuming the setting determined by CEs, the following lemma
gives a condition under which the unfounded sets w.r.t.\ $N$ for
$\tr{}{P}$ can be converted into unfounded sets w.r.t.\ $M$ for $P$.

\begin{lemma}
\label{lemma:UF-to-PUF}
Let $P$ be a program, $M$ a partial interpretation of $P$ and
$N$ a total interpretation of the program $\tr{}{P}$ such that
CE1--CE4 are satisfied.
Then, for any unfounded set $X$ for $\tr{}{P}$ w.r.t.\ $N$,
the set of atoms $Y=\sel{a\in\hb{P}}{\pot{a}\in X}$
is an unfounded set for $P$ w.r.t.\ $M$.
In addition, if $X$ is $N$-consistent,
then $Y$ is $M$-consistent.
\end{lemma}

\begin{proof}
Consider any rule $A\IF B,\naf C\in P$ such that $A\isect
Y\neq\emptyset$.  It is proven in the sequel that one of the
unfoundedness conditions \UFi--\UFiii\ applies to $A\IF B,\naf C$.
Two cases arise depending on the value of $\val{M}{B\union\naf C}$.

\begin{itemize}
\item[\bf I.]
If $\val{M}{B\union\naf C}=\false$, then
\UFi\ is directly applicable.

\item[\bf II.]
Suppose that $\val{M}{B\union\naf C}\neq\false$ which
implies $\val{N}{\pot{B}\union\naf C}=\true$ by the CEs.
Now $A\isect Y\neq\emptyset$ and the definition of $Y$ imply
$\pot{A}\isect X\neq\emptyset$.
Since $\pot{A}\IF\pot{B},\naf C\in\tr{}{P}$, $X$ is an unfounded set
for $\tr{}{P}$ w.r.t.\ $N$ and \UFi\ is not applicable to
$\pot{A}\IF\pot{B},\naf C$, we know that either \UFii\ or \UFiii\ 
applies to $\pot{A}\IF\pot{B},\naf C$.
\begin{enumerate}
\item[(i)]
If \UFii\ applies to $\pot{A}\IF\pot{B},\naf C$, then $\pot{B}\isect
X\neq\emptyset$. It follows by the definition of $Y$ that $B\isect
Y\neq\emptyset$, i.e., \UFii\ applies to $A\IF B,\naf C$.

\item[(ii)]
If \UFiii\ applies to $\pot{A}\IF\pot{B},\naf C$, then
$\val{N}{\Lor(\pot{A}-X)}=\true$. Since $\pot{A}-X=\pot{(A-Y)}$ by the
definition of $Y$, we obtain by the CEs that $\val{M}{\Lor(A-Y)}\neq\false$.
Thus \UFiii\ applies to $A\IF B,\naf C$.

\end{enumerate}
\end{itemize}

The proof of the consistency claim follows.
To establish the contrapositive of the claim, suppose that $Y$ is not
$M$-consistent. Then $Y\isect\True{M}\neq\emptyset$, i.e., there
exists an atom $a\in\hb{P}$ such that $a\in Y$ and $a\in\True{M}$.
The former implies $\pot{a}\in X$ by the definition of $Y$ while the
latter gives us $\pot{a}\in\True{N}$ by the CEs.
Thus $X\isect\True{N}\neq\emptyset$ and $X$
is not $N$-consistent.
\end{proof}

The next lemma shows that, under the specified conditions, an
unfounded set for a given disjunctive program $P$ corresponds to a
collection of unfounded sets for the translation $\tr{}{P}$.

\begin{lemma}
\label{lemma:PUF-to-UF}
Let $M$ be a partial model of a disjunctive program $P$ and 
$N$ a total interpretation of $\tr{}{P}$ satisfying the CEs. 
If $X$ is an $M$-consistent unfounded set for $P$ w.r.t.\ $M$,
then $Y=F \union U$ where $F=\sel{a,\pot{a}}{a \in X}$ and
$U\subseteq\sel{a}{a\in\False{N},\pot{a}\in\True{N}}$
is an unfounded set for $\tr{}{P}$ w.r.t.\ $N$.
\end{lemma}

\begin{proof}
Let $X$ be an $M$-consistent unfounded set for $P$ w.r.t.\ $M$ and let
$Y=F\union U$ satisfy the requirements above. Since any atom in $Y$ is
either marked or unmarked, two cases arise.

\begin{itemize}
\item[\bf I.]
Suppose that $\pot{a}\in Y$ which implies
by the definition of $Y$ that $a\in Y$.
Then it is clear that that \UFii\ applies to the consistency rule
$\pot{a}\IF a\in\tr{}{P}$.
Let us then prove that one of the unfoundedness conditions applies to
any rule $\pot{A}\IF\pot{B},\naf C\in \tr{}{P}$ satisfying
$\pot{a}\in\pot{A}$.
Since $N$ is a total interpretation, we have
$\val{N}{\pot{B}\union\naf C}=\false$ (in which case \UFi\ applies to
$\pot{A}\IF\pot{B},\naf C$) or $\val{N}{\pot{B}\union\naf C}=\true$ in
which case $\val{M}{B\union\naf C}>\false$.  Since $a\in X$ and $a\in
A$, and \UFi\ does not apply to $A\IF B,\naf C$, we only need to
consider \UFii\ and \UFiii.
If \UFii\ applies to $A\IF B,\naf C$, $\exists b \in B$ such that $b
\in X$. It follows by the definition of $Y$ that $\pot{b}\in Y$.
Hence \UFii\ applies to $\pot{A}\IF\pot{B},\naf C$.
If \UFiii\ applies to $A\IF B,\naf C$, $\exists b \in A$ such that
$\val{M}{b}>\false$ and $b\not\in X$. Then we know that
$\val{N}{\pot{b}}=\true$ by the CEs.  Further, by the definition of
$Y$, $b \not \in X$ implies $\pot{b} \not \in Y$. Hence \UFiii\
applies to $\pot{A}\IF\pot{B},\naf C$.

\item[\bf II.]
Suppose that $a \in Y$.
Then consider any rule $A\IF B,\naf\pot{C}\in \tr{}{P}$ such that
$a\in A$.  Since $N$ is a total interpretation,
$\val{N}{B\union\naf\pot{C}}=\false$ (in which case \UFi\ applies to
$A\IF B,\naf\pot{C}$) or $\val{N}{B\union\naf\pot{C}}=\true$.
In the latter case, we know that $\exists b\in A$ such that
$\val{N}{b}=\true$, since $N$ is a model of $\tr{}{P}$ by
Proposition~\ref{model-for-M-N}
(recall that $M$ is a partial model of $P$).
Then suppose that $b\in Y$, i.e., $b \in F$ or $b \in U$ by the
definition of $Y$.
If $b\in F$, then $b \in X$ by the definition of $F$. On the
other hand, $\val{N}{b}=\true$ implies $\val{M}{b}=\true$.
Thus $\True{M}\isect X\neq\emptyset$, contradicting the
$M$-consistency of $X$.
If $b\in U$, the the definition of $U$ implies $\val{N}{b}=\false$, a
contradiction.
Hence, $b\not\in Y$ and \UFiii\ applies to $A\IF B,\naf\pot{C}$.
\end{itemize}
\end{proof}

We note that the $M$-consistency of $X$ is also a necessary condition
for the correspondence to hold.

\begin{example}
Consider a disjunctive program
$
P = \set{ a \lor b \IF;\ a \IF \naf a }
$
and a partial model $M =\pair{\set{b}}{\emptyset}$ of $P$.
It can be checked easily that $X=\set{b}$ is an unfounded set for $P$
w.r.t.\ $M$: \UFiii\ applies to the only rule in which $b$ appears in
the head. But $X$ is not $M$-consistent. Now consider
\[
\begin{array}{r@{\,}c@{\,}lll@{}l}
\tr{}{P} & =\{ & a \lor b \IF;
               & \pot{a} \lor \pot{b} \IF;
               & a \IF \naf \pot{a}; \\
         &     & \pot{a} \IF \naf a;
               & \pot{a} \IF a;
               & \pot{b} \IF b & \}. \\
\end{array}
\]
The total interpretation corresponding to $M$ above is
$N=\set{\pot{a},b,\pot{b}}$.  However, $Y=\set{a,b,\pot{b}}$ is not
unfounded for $\tr{}{P}$ w.r.t.\ $N$, since for $b \in Y$ and the
first rule in $\tr{}{P}$, none of the unfoundedness conditions
applies.
\ebox
\end{example}

Let us establish Theorem \ref{theorem:PSM-iff-SM} in two separate theorems.

\begin{theorem}
\label{theorem:SM-to-PSM}
Let $P$ be a disjunctive program.
If $N$ is a stable model of the translation $\tr{}{P}$, then the
partial interpretation $M$ of $P$ satisfying the correspondence
equations CE1--CE4 is a partial stable model of $P$.
\end{theorem}

\begin{proof}
Let $N$ be a stable model of $\tr{}{P}$.
Then it follows by the presence of consistency rules
$\sel{\pot{a}\IF a}{a\in\hb{P}}$
in $\tr{}{P}$ that there is no $a\in\hb{P}$ such that $a\in N$ and
$\pot{a}\not\in N$, since $N$ is a total model of $\tr{}{P}$.
Thus it makes sense to define $M$ as the partial interpretation
satisfying CE1--CE4.  We prove that $\True{M}$ is a minimal total
model of $\GLred{P}{M}$, and $\False{M}$ is a maximal $M$-consistent
unfounded set for $P$ w.r.t. $M$.

\begin{itemize}
\item[\bf I.]
Let us first establish that for any rule $A\IF B,\naf C\in P$, $A\IF
B\in\GLred{P}{M}$ $\iff$ $A\IF B\in\GLred{\tr{}{P}}{N}$. So consider
any $A\IF B,\naf C\in P$.
It follows by the CEs and the definitions of $\GLred{P}{M}$,
$\tr{}{P}$ and $\GLred{\tr{}{P}}{N}$ that
$A\IF B\in\GLred{P}{M}$
$\iff$
there is a rule $A\IF B,\naf D\in P$ such that $D\subseteq\False{M}$
$\iff$
there is a rule $A\IF B,\naf\pot{D}\in\tr{}{P}$ such that
$\pot{D}\subseteq\False{N}$
$\iff$
$A\IF B\in\GLred{\tr{}{P}}{N}$.
Note that within these equivalences $A\IF B,\naf C$ and $A\IF B,\naf
D$ need not be the same rules of $P$.

\item[\bf II.]
Let us then prove that $\True{M}$ is a minimal total model of
$\GLred{P}{M}$. If we assume the contrary, two cases arise.

\begin{itemize}
\item
$\True{M}$ is not a total model of $\GLred{P}{M}$, i.e., there is a
rule $A\IF B\in\GLred{P}{M}$ such that $\val{\True{M}}{B}=\true$, but
$\val{\True{M}}{A}=\false$. It follows by the CEs that
$\val{N}{B}=\true$ and $\val{N}{A}=\false$. Thus $A\IF B$ is not
satisfied in $N$ and thus $N$ is not a model of $\GLred{\tr{}{P}}{N}$,
as $\GLred{P}{M}\subset\GLred{\tr{}{P}}{N}$ holds by (\textbf{I})
above. A contradiction, since $N$ is a stable model of $\tr{}{P}$.

\item
There is a total model $M'$ of $\GLred{P}{M}$ such that
$M'\subset\True{M}$. Then define a total interpretation
$N'= M'\union\sel{\pot{a}}{\pot{a} \in N}$.
By $M' \subset \True{M}$ and the CEs, we obtain $N'\subset N$
(only some unmarked atoms of $N$ are not in $N'$).
Since $M'$ is a total model of $\GLred{P}{M}$, and $M'$ and $N'$
coincide on the atoms of $\hb{P}$, every rule in $\GLred{P}{M}$ is
satisfied by $N'$.
By (\textbf{I}), the difference $\GLred{\tr{}{P}}{N}-\GLred{P}{M}$
contains only consistency rules $\pot{a}\IF a$ (for every
$a\in\hb{P}$) and rules of the form $\pot{A}\IF\pot{B}$ (for some
$A\IF B,\naf C\in P$). These rules are all satisfied by $N'$, since
$N$ is a total model of $\GLred{\tr{}{P}}{N}$, $N'\subset N$, and $N'$
and $N$ coincide on the marked atoms in $\pot{\hb{P}}$.  Thus $N'$ is
a total model of $\GLred{\tr{}{P}}{N}$.  Then $N'\subset N$ implies
that $N$ is not a minimal model of $\GLred{\tr{}{P}}{N}$ nor a total
stable model of $P$. A contradiction.
\end{itemize}

\item[\bf III.]
Since $N$ is a total stable model of $\tr{}{P}$, it holds by Theorem
\ref{theorem:SM-character} that $\False{N}$ is the greatest unfounded
set for $\tr{}{P}$ w.r.t.\ $N$. Moreover, $\False{N}$ is
$N$-consistent, since $\False{N}\isect\True{N}=\emptyset$.
Note that $\pot{a}\in\False{N}$ implies $a\in\False{N}$,
since $N$ satisfies $\pot{a}\IF a\in\tr{}{P}$. Thus
$\False{M}=
 \sel{a\in\hb{P}}{a\in\False{N}\mbox{ and }\pot{a}\in\False{N}}=
 \sel{a\in\hb{P}}{\pot{a}\in\False{N}}$.
It follows by Lemma~\ref{lemma:UF-to-PUF} that $\False{M}$ is an
$M$-consistent unfounded set for $P$ w.r.t.\ $M$.

Then assume that $\False{M}$ is not maximal, i.e., there is an
$M$-consistent unfounded set $X$ for $P$ w.r.t.\ $M$ such that $X
\supset\False{M}$. So there is an atom $a\in X$ such that $a\not \in
\False{M}$. Then $a\not\in\False{M}$ implies $a\in\True{M}$ or $a
\in\Undef{M}$. In both cases, by the CEs, $\pot{a}\in\True{N}$,
i.e., $\pot{a}\not\in\False{N}$.
Then construct $Y=\sel{a,\pot{a}}{a\in X}$. According to
Lemma~\ref{lemma:PUF-to-UF}, that $X$ is an $M$-consistent
unfounded set for $P$ w.r.t.\ $M$ implies that $Y$ is an unfounded set
for $\tr{}{P}$ w.r.t $N$.  However, $a\in X$ implies $\pot{a}\in Y$
but $\pot{a}\not\in \False{N}$.  Thus $\pot{a}\in\True{N}$ indicating
that $N$ is not unfounded free for $\tr{}{P}$. Consequently, $N$ is
not a stable model of $\tr{}{P}$ by the characterization of stable
models in Theorem \ref{theorem:SM-character}, a contradiction.

\end{itemize}
\end{proof}

\begin{theorem}
\label{theorem:PSM-to-SM}
Let $P$ be a disjunctive program.
If $M$ is a partial stable model of $P$, then the total interpretation
$N$ satisfying the correspondence equations CE1--CE4 is a stable model
of the translation $\tr{}{P}$.
\end{theorem}

\begin{proof}
Suppose that $M$ is a partial stable model of $P$. Then we know by
Theorem \ref{theorem:PSM-character} that (i) $\True{M}$ is a minimal
total model of $\GLred{P}{M}$ and (ii) $\False{M}$ is a maximal
$M$-consistent unfounded set for $P$ w.r.t.\ $M$.
Then define $N$ as the total interpretation of $\tr{}{P}$ satisfying
the CEs. It follows by Lemma \ref{lemma:PUF-to-UF} that
$\False{N}=\sel{a,\pot{a}}{a\in\False{M}}\union\Undef{M}$ is an
unfounded set for $\tr{}{P}$ w.r.t.\ $N$.

Let us then assume that $N$ is not a stable model of $\tr{}{P}$.
Equivalently, it holds by Theorem \ref{theorem:SM-character} that
$\False{N}$ is not the greatest unfounded set for $\tr{}{P}$
w.r.t. $N$. So there is an unfounded set $X$ for $\tr{}{P}$ w.r.t.\
$N$ such that $\False{N}\subset X$ and $\True{N}\isect X\neq\emptyset$
hold. It follows by Proposition~\ref{no-effect} that $X$ is also an
unfounded set for $\GLred{\tr{}{P}}{N}$ w.r.t. $N$.

Then consider any $A\IF B\in\GLred{P}{M}$ for which there is a rule
$A\IF B,\naf C\in P$ such that $C\subseteq\False{M}$. It follows by
the CEs that $\pot{C}\subseteq\False{N}$. Since $A\IF
B,\naf\pot{C}\in\tr{}{P}$, it follows that $A\IF
B\in\GLred{\tr{}{P}}{N}$. Thus
$\GLred{P}{M}\subset\GLred{\tr{}{P}}{N}$ holds, as
$\GLred{\tr{}{P}}{N}$ contains among others the consistency rules
$\sel{\pot{a}\IF a}{a\in\hb{P}}$.

Recall that $\True{M}=\True{N}\isect\hb{P}$ is a minimal total model
of $\GLred{P}{M}$. We also distinguish a set of atoms
$X'=X\isect\hb{P}$. Let us then establish that $X'$ is an unfounded
set for $\GLred{P}{M}$ with respect to $\True{M}$ in the two-valued sense.
\begin{itemize}
\item[{\bf I.}]
If $X'$ is not such a set, it follows by Definition
\ref{def:unfounded-sets} that there is $A\IF B\in\GLred{P}{M}$ with
$A\isect X'\neq\emptyset$ such that $B\subseteq\True{M}$, $B\isect
X'=\emptyset$ and $(A-X')\isect\True{M}=\emptyset$.
It follows that $A\IF B\in\GLred{\tr{}{P}}{N}$, as
$\GLred{P}{M}\subset\GLred{\tr{}{P}}{N}$. Since $A$ and $B$ are
subsets of $\hb{P}$, $\True{M}=\True{N}\isect\hb{P}$ and
$X'=X\isect\hb{P}$, we obtain $A\isect X\neq\emptyset$,
$B\subseteq\True{N}$, $B\isect X=\emptyset$ and
$(A-X)\isect\True{N}=\emptyset$. Then $X$ is not
an unfounded set for $\GLred{\tr{}{P}}{N}$ w.r.t.\ $N$,
a contradiction.
\end{itemize}
It follows by Lemma \ref{lemma:set-unfounded-false} that $\True{M}-X'$
is a total model of $\GLred{P}{M}$. It follows by the minimality of
$\True{M}$ that $\True{M}\isect X'=\emptyset$ and $\True{M}\isect
X=\emptyset$.
Moreover, it follows by Lemma~\ref{lemma:UF-to-PUF} that
$Y=\sel{a\in\hb{P}}{\pot{a}\in X}$ is an unfounded set for $P$ w.r.t.\
$M$. It remains to establish that $Y$ is $M$-consistent and
$\False{M}\subset Y$.
\begin{itemize}
\item[\bf II.]
Suppose that $Y$ is not $M$-consistent, i.e., it holds for some
$\at{a}\in\hb{P}$ that (a) $a\in Y$ and (b) $a\in\True{M}$.
Now (b) implies by the CEs that $a\in\True{N}$ and
$\pot{a}\in\True{N}$.
On the other hand, it follows by (a) and the definition of $Y$ that
$\pot{a}\in X$. Thus one of the unfoundedness conditions applies to
the rule $\pot{a}\IF a\in\tr{}{P}$, as $X$ is an unfounded set for
$\tr{}{P}$ w.r.t.\ $N$.
Now \UFi\ is not applicable, as $\at{a}\not\in\False{N}$, and \UFiii\
is not applicable, as $\pot{a}\in X$.  Thus \UFii\ must be applicable
to $\pot{a}\IF a$. It follows that $\at{a}\in X$, too.  Then there is
$a\in\hb{P}$ such that $a\in\True{M}$ and $a\in X$. A contradiction
with $\True{M}\isect X=\emptyset$ established above.

\item[\bf III.]
Consider any $a\in\False{M}$.  Thus $a\in\hb{P}$ and
$\pot{a}\in\False{N}$ follows by the CEs.  Then $\False{N}\subset X$
implies $\pot{a}\in X$ as well as $a\in Y$ by the definition of $Y$.
Thus $\False{M}\subseteq Y$.
On the other hand, recall that $\True{N}\isect X\neq\emptyset$ and
$\True{M}\isect X=\emptyset$. Then $\pot{a}\in \True{N}\isect X$ holds
for some $a\in\hb{P}$. It follows that $\pot{a}\in\True{N}$ and
$\pot{a}\in X$.  The former implies $\at{a}\not\in\False{M}$ by the
CEs.  The latter implies $a\in Y$ by the definition of $Y$. Hence
$\False{M}\subset Y$.
\end{itemize}
Thus $\False{M}$ is not a maximal $M$-consistent unfounded set for $P$
w.r.t.\ $M$, a contradiction.
Hence $N$ must be a stable model of $\tr{}{P}$.
\end{proof}

It is worthwhile at this point to briefly comment on the proof of
Theorem~\ref{theorem:PSM-iff-SM} as given in \cite{SMR97:ilps}, which
proceeds in several steps. Given two partial interpretations
$M$ and $M'$ of a disjunctive program $P$, let $N$ and $N'$,
respectively, be the corresponding total interpretations of
$\tr{}{P}$ such that CE1--CE4 are satisfied.
Firstly, it can be shown that $M$ is a partial model of $\red{P}{M'}$
if and only if $N$ is a total model of $\GLred{\tr{}{P}}{N'}$.
Secondly, since the truth-ordering for partial
interpretations corresponds to the subset ordering for total
interpretations, it can be shown that
the minimal partial models of $\red{P}{M'}$ correspond to
the minimal total models of $\GLred{\tr{}{P}}{N'}$.
Thirdly, based on a characterization of partial models in general
\cite{SMR97:ilps}, we conclude that $M$ is a partial stable model of
$P$ if and only if $N$ is a total stable model of $\tr{}{P}$.

Looking back to results established so far, we know by Theorem
\ref{theorem:PSM-to-SM} that any partial stable model $M$ of $P$ can
be mapped to a stable model
\begin{equation}
\label{eq:expand}
f(M)=\True{M}\union\pot{(\True{M}\union\Undef{M})}
\end{equation}
of $\tr{}{P}$. Similarly, any stable model $N$ of $\tr{}{P}$
can be projected to a partial stable model
\begin{equation}
\label{eq:project}
g(N)=
 \pair{\sel{a\in\hb{P}}{a\in N}}
      {\sel{a\in\hb{P}}{a\not\in N\mbox{ and }\pot{a}\not\in N}}
\end{equation}
of $P$ by Theorem \ref{theorem:SM-to-PSM}. These equations and the
corresponding theorems indicate that $f$ and $g$ are functions between
the set of partial stable models of $P$ and the set of stable models
of $\tr{}{P}$. In the sequel, it is established that these functions
are bijections, which means that our translation technique does not
yield spurious models for programs although new atoms are used. This
is highly desirable from the knowledge representation perspective.

\begin{theorem}
\label{theorem:one-to-one-correspondece}
The partial stable models of a disjunctive program $P$
and the total stable models of the translation $\tr{}{P}$
are in a one-to-one correspondence.
\end{theorem}

\begin{proof}
Let $f$ and $g$ be defined by the equations (\ref{eq:expand}) and
(\ref{eq:project}), respectively.  It is straightforward to see that
$f$ is injective, i.e., $f(M_1)=f(M_2)$ implies $M_1=M_2$.
Then assume that $g(N_1)=g(N_2)$ holds for some stable models $N_1$
and $N_2$ of $\tr{}{P}$. It follows by the definition of $g$ for
any $a\in\hb{P}$ that (i) $a\in N_1$ $\iff$ $a\in N_2$ and (ii)
$a\not\in N_1$ and $\pot{a}\not\in N_1$ $\iff$ $a\not\in N_2$ and
$\pot{a}\not\in N_2$.
Then consider any $a\in\hb{P}$ such that $\pot{a}\in N_1$. Two cases
arise.  If $a\in N_1$, it follows by (i) that $a\in N_2$. Since $N_2$
satisfies the rule $\pot{a}\IF a\in\GLred{\tr{}{P}}{N_2}$, we obtain
$\pot{a}\in N_2$. On the other hand, if $a\not\in N_1$ it follows by
(i) that $a\not\in N_2$. Assuming that $\pot{a}\not\in N_2$ implies by
(ii) that $\pot{a}\not\in N_1$, a contradiction. Hence $\pot{a}\in
N_2$ also in this case. By symmetry, $\pot{a}\in N_2$ implies
$\pot{a}\in N_1$.

Thus it holds for any $a\in\hb{P}$ that (iii) $\pot{a}\in N_1$ $\iff$
$\pot{a}\in N_2$. It follows by (i) and (iii) that $N_1=N_2$ so that
$g$ is injective, too. Thus $f$ and $g$ are bijections and
inverses of each other, as $g(f(M))=M$ and $f(g(N))=N$ hold
for any (partial) stable models $M$ and $N$. Hence the claim.
\end{proof}

None of the preceding proofs relies on the assumption that the given
program is finite. Therefore, all of these results presented in this
section apply in the non-ground case as well.


\subsection{QUERY ANSWERING}

Let us yet address the possibility of using an inference engine for
computing total stable models to answer queries concerning partial
stable models. This is highly interesting, because there are already
systems available for computing total stable models 
\cite{CMODELS,LZ02:aaai,DLV,SNS02:aij}
while partial stable models lack
implementations. Here we must remind the reader that partial stable
models can be used in different ways in order to evaluate queries.
Typically two modes of reasoning are used: {\em certainty inference}
and {\em possibility inference}. In the former approach, a query $Q$
should be true in all (intended) models of $P$ while $Q$ should be
true in some (intended) model of $P$ in the latter approach. Moreover,
maximal partial stable models (under set inclusion) are sometimes
distinguished; this is how {\em regular models} and {\em preferred
extensions} are obtained for normal
programs~\cite{Dung95:jlp,Sacca90:pods,YY94:jcss}.
We are particularly interested in possibility inference where the
maximality condition makes no difference (see
\cite{ELS98:tcs,Sacca97:jcss} for certainty inference):
$\val{M}{Q}=\true$ for some partial stable model $M$ of $P$ if and only if
$\val{M'}{Q}=\true$ for some maximal partial stable model $M'$ of $P$.

We consider queries $Q$ that are sets of literals over $\hb{P}$ and
queries are translated in harmony with the CEs:
$\tr{}{Q}=Q\union\pot{Q}$. As a direct consequence of Theorem
\ref{theorem:PSM-to-SM} and CE1, we obtain the following.
\begin{corollary}
A query $Q$ is true in a (maximal) partial stable model
of $P$ if and only if $\tr{}{Q}$ is true in a stable model of $\tr{}{P}$.
\end{corollary}

What about using a query answering procedure for partial stable models
to answer queries concerning stable models?  A slight extension of the
translation $\tr{}{P}$ is needed for this purpose: let $\tr{2}{P}$ be
$\tr{}{P}$ augmented with a set of rules $\sel{f\IF\pot{a},\naf
a}{a\in\hb{P}}$ where $f\not\in\hb{P}$ is a new atom. The purpose of
these additional rules is to detect partial stable models with
remaining undefined atoms.  A query $Q$ is translated into
$\tr{2}{Q}=Q\union\set{\naf f}$.
\begin{corollary}
A query $Q$ is true in a stable model of $P$ if and only if
$\tr{2}{Q}$ is true in a partial stable model of $\tr{2}{P}$.
\end{corollary}

This result allows query answering for stable models to be conducted
by a procedure for partial stable models, e.g., by the abductive
procedure of Eshghi and Kowalski \cite{EK89:iclp}.
 

\section{UNFOLDING DISJUNCTIONS}
\label{section:disjunctions}

In this section we develop a method for reducing the task of computing
a (total) stable model of a disjunctive program to computing stable
models for normal (disjunction-free) programs. This objective demands
us to {\em unfold}\,\footnote{The idea of unfolding disjunction
generally refers to performing some transformations on disjunctions in
order to remove them \cite{BD99:jlp,DGM96:fi,SS97:jlp}. However, such
transformations do not necessarily remove all disjunctions or do not
preserve stable semantics.} disjunctions from programs in a way or
another.
Since the problem of deciding whether a disjunctive program has a
stable model is $\Sigma^p_2$-complete~\cite{EG95:amai} whereas the
problem is $\NP$-complete in the non-disjunctive
case~\cite{MT91:jacm}, the reduction cannot be computable in
polynomial time unless the polynomial hierarchy collapses. This is why
our reduction is based on a generate and test approach.

The basic idea is that given a disjunctive program $P$ we compute its
stable models in two phases: (i) we \emph{generate} model candidates
and (ii) \emph{test} candidates for stability until we find a suitable
model.  For generating model candidates we construct a normal program
$\gen{P}$ such that the stable models of $\gen{P}$ give the candidate
models. For testing a candidate model $M$ we build another normal
program $\test{P}{M}$ such that $\test{P}{M}$ has no stable models 
if and only if
$M$ is a stable model of the original disjunctive program $P$.
Hence, given a procedure for computing stable models for normal
programs all stable models of a disjunctive program $P$ can be
generated as follows: for each stable model $M$ of $\gen{P}$, decide
whether $\test{P}{M}$ has a stable model and if this is not the case,
output $M$ as a stable model of $P$.
This kind of a generate and test approach is used also in \sys{dlv}~\cite{DLV}
which is a state-of-the-art system for disjunctive programs. 
The difference is that we reduce the test and generate subtasks directly
to problems of computing stable models of normal programs whereas 
in \sys{dlv} special techniques for the two subtasks have been
developed based on the notion of unfounded sets for disjunctive
programs. 

It is easy to construct a normal program for generating
candidate models for a disjunctive program $P$.
Consider, e.g., a program $\genO{P}$ which contains
for each atom $a \in \hb{P}$, two rules $a \IF \naf \hat{a}\END \hat{a}
\IF \naf a$ where $\hat{a}$ is a new atom
denoting the complement of the atom $a$, i.e., $a$ is in a stable model
exactly when $\hat{a}$ is not.
These rules generate stable models
corresponding to every subset of $\hb{P}$. In order to prune this set
of models to those with all rules in $P$ satisfied, it is sufficient
to include a rule
\begin{equation}
f \IF \naf f, \naf a_1,\ldots,\naf a_k,b_1,\ldots,
b_m, \naf c_1,\ldots, \naf c_n
\label{eq:consistency}
\end{equation}
for each rule of the form~(\ref{eq:drule}) in $P$ where $f$ is a new
atom.  As $f$ cannot be in any stable model, the rule functions as an
integrity constraint eliminating the models where each $b_i$ is
included, every $c_j$ is excluded but no $a_l$ is included.

In order to guarantee completeness, it is sufficient that for each
stable model $M$ of $P$ there is a corresponding model candidate which
agrees with $M$ w.r.t.\ $\hb{P}$.  It is clear that $\genO{P}$
satisfies this condition.  However, for efficiency
it is important to devise a generating program that has as few as
possible (candidate) stable models provided that completeness is not
lost. 
An obvious shortcoming of $\genO{P}$ is that it generates many
candidates even if the program $P$ is disjunction-free.
In order to solve this problem 
we construct 
for given a disjunctive program $P$ a generating program $\genI{P}$ as
follows:  
\begin{eqnarray*}
\genI{P} & = & 
\begin{array}[t]{@{}l@{}}
\{a \IF \naf \hat{a}, B, \naf C \mid A \IF B, \naf C \in \dlp{P}, a
                                   \in A \} \union \mbox{} \\
\{\hat{a}\IF \naf a \mid a \in \heads{\dlp{P}} \} \union \mbox{} \\
\{f \IF \naf f, \naf A, B, \naf C \mid 
                                    A \IF B, \naf C \in \dlp{P} \} \cup
                                   \nlp{P} 
\end{array} 
\end{eqnarray*}
where $\nlp{P}$ is the set of the normal rules in $P$ and $\dlp{P}$ are
the other 
(proper disjunctive) rules in $P$, i.e.\ $P = \nlp{P} \cup \dlp{P}$, and
$\heads{\dlp{P}}$ is the set of atoms appearing in the heads of the rules in
$\dlp{P}$. 

The program $\genI{P}$ has typically far fewer stable models than
$\genO{P}$ and the number of ``extra'' candidate models which do not
match stable models of $P$ is related to the number of
disjunctions in $P$. 
For example, if $P$ is disjunction-free, 
the stable models of $\genI{P}$ correspond exactly to the stable
models of $P$.
However, for a disjunctive program $P$, $\genI{P}$ can easily have 
``extra'' stable models.
Consider, e.g.,  
\begin{equation}
P = \{ a \lor b \IF \}
\label{eq:disjunction}
\end{equation}
for which 
$\genI{P} = \{
a \IF \naf \hat{a}\END \hat{a} \IF \naf{a}\END 
b \IF \naf \hat{b}\END \hat{b} \IF \naf{b}\END
f \IF \naf f, \naf a, \naf b\} $ 
has a stable model, $\{a,b\}$, not corresponding to a stable model of
$P$.  
In fact, $\genI{P}$ only requires for each proper disjunctive rule in $P$
that some non-empty subset of the head atoms of the rule is included in
the model candidate when the body of the rule holds. Hence, for such a rule
with $d$ disjuncts in the head there are $2^d -1$
possible subsets and in the worst case $2^d - 2$ of these could lead to 
``extra'' model candidates.
This means that in the worst case $\genI{P}$ can have an exponential
number of ``extra'' model candidates w.r.t.\ the number of disjunctions
in $P$.

In order to decrease the number of ``extra'' models we introduce
a technique exploiting a key property of {\em supported models}
\cite{BD97:jlp}: each atom $a$ true in a model $M$ of $P$ must have
a rule {\em supporting} it, i.e., there is a rule $A\IF B,\naf C\in P$
such that $a\in A$, $\val{M}{B\union\naf C}=\true$, and
$\val{M}{\Lor(A-\set{a})}=\false$. 
Since every stable model of $P$ is also a supported model of $P$, it
makes perfect sense to require supportedness from the candidate
stable models.
For this, we introduce a new atom $\supp{a}$, which denotes the fact
that atom $a$ has a supporting rule, for each atom $a$ appearing in the
head of a disjunctive rule. The
intuition behind the set of rules $\support{P}$ below is that a rule can
support exactly one of its head atoms and we may exclude every model
that has an atom without a supporting rule:
\begin{multline}
\support{P} = \\
\{\supp{a} \IF \naf (A-\{a\}), B, \naf C  \mid A \IF B, \naf C \in P, a
                                   \in A \isect \heads{\dlp{P}}\}
\ \union \\
\{f \IF \naf f, a, \naf \supp{a} \mid a \in \heads{\dlp{P}} \}
\end{multline}
where $\heads{\dlp{P}}$ is the set of atoms appearing in the heads of
the proper disjunctive rules in $P$. 
For example, for $P$ in (\ref{eq:disjunction}), 
\begin{center}
$\support{P} = \{\supp{a} \IF \naf b\END \supp{b} \IF \naf a\END  
f \IF \naf f, a, \naf \supp{a}\END 
f \IF \naf f, b, \naf \supp{b} \}$.
\end{center}
Now $\genI{P} \cup \support{P}$ has
exactly two stable models $\{a,\supp{a}, \hat{b}\}$ and $\{b, \supp{b}, \hat{a}\}$
corresponding to the two stable models $\{a\}$ and
$\{b\}$ of $P$.

Combining this idea with $\genI{P}$ gives a promising generating program 
\begin{equation}
\gen{P} = \genI{P} \cup \support{P} 
\label{eq:gen}
\end{equation}
which still preserves completeness. 

\begin{proposition}
Let $P$ be a disjunctive program. Then if $M$ is a stable model
of $P$, there is a stable model $N$ of $\gen{P} = \genI{P} \cup \support{P}$
with $M=N\isect\hb{P}$.
\end{proposition}

\begin{proof}
Let $M$ be a stable model of $P$ and
\[N=M \union \sel{\hat{a}}{a\in\heads{\dlp{P}}-M} \union 
  \sel{\supp{a}}{a\in M\isect \heads{\dlp{P}}}.\]
Now clearly $M=N\isect\hb{P}$.  
We show first that (i) $N$ is a model
of $\GLred{\gen{P}}{N}$ and then that 
(ii) if there is a model $N'$ of $\GLred{\gen{P}}{N}$ such that $N' \subseteq N$
then $N \subseteq N'$ holds. 
These together imply that $N$ is a stable model of $\gen{P}$. 

For property~(i) consider rules in 
$\GLred{\gen{P}}{N} = \GLred{\genI{P}}{N} \cup \GLred{\support{P}}{N}$
starting with those in $\GLred{\genI{P}}{N}$. 
Suppose $a \IF B \in \GLred{\genI{P}}{N}$. If $a \in \heads{\dlp{P}}$, 
then $\hat{a} \not\in N$ and, hence,
$a \in M \subseteq N$. Otherwise if $a \not\in \heads{\dlp{P}}$, then 
$a \IF B \in \GLred{P}{M}$ implying that $N (\supseteq M)$ satisfies
$a \IF B$. 
If $\hat{a} \IF \in \GLred{\genI{P}}{N}$, then $a \not\in M$ and 
$\hat{a} \in N$. 
If $f \IF B \in \GLred{\genI{P}}{N}$, then there is a rule 
$ A \IF B, \naf C \in \dlp{P}$ such that 
$A \isect M = \emptyset$ and $A \IF B \in \GLred{P}{M}$. 
As $M$ is a model of $\GLred{P}{M}$, $B
\not\subseteq M$ and, consequently, $B \not\subseteq N$. 
Thus, $N$ is a model of $\GLred{\genI{P}}{N}$.
Next consider rules in $\GLred{\support{P}}{N}$. 
If $\supp{a} \IF B \in \GLred{\support{P}}{N}$, then there is a rule 
$ A \IF B, \naf C \in P$ such that 
$A \IF B \in \GLred{P}{M}$ and $(A - \{a\}) \isect M = \emptyset$.
If $B \subseteq N$, then $B \subseteq M$ and, hence, $A \isect M \not=
\emptyset$ as $M$ is a model of $\GLred{P}{M}$. Thus, $a \in M$.
If $ f \IF a \in \GLred{\support{P}}{N}$, then $\supp{a} \not\in N$, and $a
\not \in N$. This implies that $N$ is a model of $\GLred{\support{P}}{N}$
and that (i) holds.

For property~(ii) consider a model $N'$ of $\GLred{\gen{P}}{N}$ such
that $N' \subseteq N$. 
First we show that $N'\isect \hb{P}$ is a model of $\GLred{P}{M}$
implying that $ M \subseteq N'\isect \hb{P}$. 
Consider $A \IF B \in \GLred{P}{M}$.  If the body $B$ is true in
$N'\isect \hb{P} \subseteq N \isect \hb{P} = M$,
then at least one $a\in
A\isect M$. Then $a\IF B \in \GLred{\gen{P}}{N}$ and $a\in
N'$. Hence, $M \subseteq N'\isect \hb{P}$.
If $\hat{a} \in N$, then $a \not\in N$ and, hence,
$\hat{a} \IF \in \GLred{\genI{P}}{N}$ implying $\hat{a} \in N'$. 
If $\supp{a} \in N$, then $a \in M\isect \heads{\dlp{P}}$. Then there is a rule 
$A \IF B \in \GLred{P}{M}$ such that $B \subseteq M$ but $(A -\{a\})
\isect M = \emptyset$. This is because otherwise $M - \{a\}$ would be a
model of $\GLred{P}{M}$ contradicting the minimality of $M$.
Hence, $\supp{a} \IF B \in \GLred{\support{P}}{N}$ and $B \subseteq M
\subseteq N'$ implying $\supp{a} \in N'$. Thus, 
$N \subseteq N'$ and (ii) holds.
\end{proof}

A (total) model candidate $M \subseteq \hb{P}$ is a stable model of a
program $P$ if it is a minimal model of the GL-transform $\GLred{P}{M}$ of the
program.  This test can be reduced to an unsatisfiability problem in
propositional logic using techniques presented
in~\cite{Niemela96:tab}: $M$ is a minimal model of $P^M$ if and only if
\begin{equation}
P^M \union  \{\neg a \mid a \in \hb{P}-M\}
\union 
\{\neg b_1 \lor \cdots\lor \neg b_m\}
\label{eq:min-test}
\end{equation}
is unsatisfiable where $M = \{ b_1 ,\ldots, b_m\}$ and the rules in
$P^M$ are seen as clauses.  This can be determined by testing
non-existence of stable models for a normal program $\test{P}{M}$
which is constructed for a disjunctive program $P$ and a total
interpretation $M \subseteq \hb{P}$ as follows:
\[
\begin{array}{@{}l@{}}
\test{P}{M} =
\begin{array}[t]{l@{}}
\{a \IF \naf\hat{a}, B  \mid \begin{array}[t]{@{}l@{}}
                                  A \IF B \in \dlp{P}^M, 
                                  a \in A \isect M, B \subseteq M \} \union \mbox{} 
                                   \end{array}
\\
\{\hat{a}\IF \naf a \mid a \in \heads{\dlp{P}} \} \union \mbox{} \\
\{f \IF \naf f, \naf A, B \mid \begin{array}[t]{@{}l@{}}
                                    A \IF B \in \dlp{P}^M,  
                                    B \subseteq M \} \union \mbox{} 
                                   \end{array}
\\ 
\{a \IF B \in \nlp{P}^M  \mid \begin{array}[t]{@{}l@{}}
                                  a \in M, B \subseteq M \} \union \mbox{} 
                                   \end{array}
\\
\{f \IF \naf f, M \}
\end{array} 
\end{array} 
\]
where $\nlp{P}$ is the set of the normal rules in $P$ and $\dlp{P}$ are
the proper disjunctive rules in $P$ and
$\heads{\dlp{P}}$ is the set of atoms appearing in the heads of the rules in
$\dlp{P}$. 
The idea is that stable models of $\test{P}{M}$ capture models of the reduct
$P^M$ that are properly included in $M$. 

\begin{proposition}
\label{prop:mintest}
Let $P$ be a disjunctive program and $M$ a (total) model of $P$. 
Then $M$ is a minimal model of $P^M$ 
if and only if 
$\test{P}{M}$ has no stable model.
\end{proposition}

\begin{proof}
Let $M \subseteq \hb{P}$ be a total model of $P$.

($\Rightarrow$)
Let $N$ be a stable model of $\test{P}{M}$. If $a\in\hb{P}-M$, then
there is no rule with $a$ in the head in $\test{P}{M}$ and $a \not\in N$.
Hence, $N\isect\hb{P}\subseteq M$. As $f\not\in N$ and
$f\IF M \in\GLred{\test{P}{M}}{N}$, there is some $a\in M$ such
that $a\not\in N\isect\hb{P}$.
Consider $A \IF B \in \GLred{P}{M}$.
Let $B\subseteq N\isect\hb{P}\subseteq M$ but suppose $A\isect
N\isect\hb{P} =\emptyset$. 
If $A = \{a\}$, $a \IF B\in\GLred{\test{P}{M}}{N}$ and $a \in N$,  a
contradiction. Otherwise $f\IF B\in\GLred{\test{P}{M}}{N}$ and
$f\in N$, a contradiction.  Hence, $N\isect\hb{P}$ is a model of
$\GLred{P}{M}$ but $N\isect\hb{P}\subset M$ implying that $M$ is not a
minimal model of $\GLred{P}{M}$.

($\Leftarrow$)
Assume that $M$ is not a minimal model of $\GLred{P}{M}$.  As $M$ is a
model of $\GLred{P}{M}$, there is a minimal model $M'\subset M$ of
$\GLred{P}{M}$.
We show that
$N=M'\union\sel{\hat{a}}{a\in\heads{\dlp{P}}-M'}$
is a minimal model of $\GLred{\test{P}{M}}{N}$,
i.e., $N$ a stable model of $\test{P}{M}$. Now
$\GLred{\test{P}{M}}{N}=$
\[
\begin{array}[t]{@{}l@{}}
\{a \IF B  \mid \begin{array}[t]{@{}l@{}}
                          A \IF B \in \dlp{P}^M,
                          a \in A \isect M \isect M', B \subseteq M \} \union \mbox{} 
                                   \end{array}
\\
\{\hat{a}\IF \mid a \in \heads{\dlp{P}}-M' \} 
\union \mbox{} \\
\{f \IF B \mid \begin{array}[t]{@{}l@{}}
                                    A \IF B \in \dlp{P}^M,  A \isect M' = \emptyset, 
                                    B \subseteq M \} \union \mbox{} 
                                   \end{array}
\\ 
\{a \IF B \in \nlp{P}^M \mid \begin{array}[t]{@{}l@{}}
                          a \in M, B \subseteq M \} \union \mbox{} 
                                   \end{array}
\\
\{f \IF M  \} .
\end{array} 
\]
It is easy to check that (i)~$N$ is a model of
$\GLred{\test{P}{M}}{N}$. 
Assume there is a model $N'$ of $\GLred{\test{P}{M}}{N}$
such that $N' \subseteq N$ holds.
We show that then (ii)~$N \subseteq N'$ holds
as follows. We notice that for all $a \in
\hb{P}$, $a \in N'$ implies $a\in M'$.
Consider $A \IF B \in \GLred{P}{M}$.
If $B$ is true in $N'\isect\hb{P}$, then $B$ is true in $M'$ and, thus, 
$B \subseteq M$ and some $a \in A\isect M\isect M'$.  Then $a \IF B \in
\GLred{\test{P}{M}}{N}$ and $a \in N'$. Hence, $N'\isect\hb{P}$ is a
model of $\GLred{P}{M}$ which implies $M' \subseteq N'\isect\hb{P}$. 
If $\hat{a} \in N$, then $a \not\in M'$ and $\hat{a} \IF
\in \GLred{\test{P}{M}}{N}$ implying $\hat{a} \in N'$. 
Then $N \subseteq N'$ holds.  
Now (i) and (ii) imply that $N$ is a minimal model
of $\GLred{\test{P}{M}}{N}$ and, hence,  a
stable model of $\test{P}{M}$.
\end{proof}

\begin{example}
Consider a disjunctive program $P$ and its generator $\gen{P}$:
\[
\begin{array}{l}
P = 
\begin{array}[t]{l}
\{a \lor b \IF \naf c\}
\end{array}
\\
\\
\gen{P}= \{ 
\begin{array}[t]{@{\,}l}
a \IF \naf \hat{a},\naf c\END 
b \IF \naf \hat{b},\naf c\END  \\
\hat{a} \IF \naf a\END \hat{b} \IF \naf b\END
\\
f \IF \naf f, \naf a, \naf b, \naf c \END \\
\supp{a} \IF \naf b, \naf c \END
\supp{b} \IF \naf a, \naf c \END \\
f \IF \naf f,  a, \naf \supp{a} \END
f \IF \naf f,  b, \naf \supp{b} 
\; \}
\end{array}
\end{array}
\]
For a stable model $\{b, \supp{b}, \hat{a}\}$ of
$\gen{P}$ the corresponding model candidate is $M_1=\{b, \supp{b}, \hat{a}\}
\isect \hb{P} = \{b\} $ and the test program:
\[
\begin{array}[t]{l@{\,}l}
\test{P}{M_1} = \{ 
& b \IF \naf \hat{b}\END \\
& \hat{a} \IF \naf a\END  \hat{b} \IF \naf b\END \\
& f \IF \naf f, \naf a, \naf b \END 
f \IF \naf f, b 
\; \}
\end{array}
\]
$\test{P}{M_1}$ has no stable models and, hence, $M_1$ is a stable model
of $P$. 
\ebox
\end{example}
The simple generate and test paradigm can be optimized by building
model candidates gradually. This means that we start from the empty
partial interpretation and extend the interpretation step by step. An
interesting observation is that the technique for testing minimality
can be used to rule out a partial model candidate of $\gen{P}$ at any
stage of the search and not just when a total model of the program $P$
has been found. This can be done by treating a partial interpretation
$M$ as a total interpretation where undefined atoms are taken to be
false and using the $\test{P}{M}$ program.

\begin{proposition}
\label{prop:early-test}
Let $P$ be a disjunctive program and $M$ a total interpretation.  If
$\test{P}{M}$ has a stable model, then there is no (total) stable
model $M'$ of $P$ such that $M \subseteq M'$.
\end{proposition}

\begin{proof}
Let $\test{P}{M}$ have a stable model.  As shown in the proof of
Proposition~\ref{prop:mintest}, then there is a model $M''$ of
$\GLred{P}{M}$ with $M'' \subset M$.
Consider any total interpretation $M'$ such that $M \subseteq M'$ and
$M'$ is a model of $\GLred{P}{M'}$.
Now $M'$ is not a minimal model of $\GLred{P}{M'}$ as
$\GLred{P}{M'}\subseteq \GLred{P}{M}$ and, hence, $M''$ is a model of
$\GLred{P}{M'}$ but $M'' \subset M \subseteq M'$.
\end{proof}

Notice that for a total interpretation $M$,
Proposition~\ref{prop:early-test} can only be used for eliminating
stable models of $P$ extending $M$. For guaranteeing the existence of
a stable model of $P$, a total model of $P$ needs to be found making
Proposition~\ref{prop:mintest} applicable.

Our approach to testing minimality of model candidates differs from
that used in \sys{dlv}~\cite{KLP03:aij}. We check minimality by
directly searching for models of the reduct strictly contained in the
candidate model. In \sys{dlv} a dual approach is used based on the
notion of unfounded sets for disjunctive programs~\cite{LRS97:ic} and minimality
testing is done using a SAT solver. Our approach could be implemented
straightforwardly using a SAT solver, too, but we have chosen to use
the same logic program core engine for generating and testing subtasks
in order to keep the implementation as simple as possible.  A basic
difference is that in our approach the set of clauses
(\ref{eq:min-test}) used for minimality testing follow the structure
of the original program whereas in the \sys{dlv} approach dual clauses
(with each literal complemented) are employed.  Moreover, \sys{dlv}
employs a couple of optimizations which have not been exploited in our
approach. First, \sys{dlv} adopts specialized algorithms for some
syntactically recognizable classes of rules like head cycle free
programs. Second, \sys{dlv} employs modular evaluation techniques for
minimality testing where the program is divided into components based
on its dependency graph and the minimality of a candidate model is
tested for each component separately by exploiting specialized
algorithms for components with corresponding restricted form whenever
possible. For a more detailed comparison, see~\cite{KLP03:aij}.


\section{EXPERIMENTS}
\label{section:experiments}

In this section, we compare \sys{dlv}~\cite{DLV}, a state-of-the-art
implementation of the stable model semantics for disjunctive logic
programs, with an implementation of the generate and test approach of
the previous section which we call \sys{GnT}. 
In Section~\ref{section:implementation} we explain briefly implementation
techniques employed in \sys{GnT} and 
explain the setup for the experiments. 
For comparisons we use three families of test problems 
related to reasoning about {\em minimal models}~\cite{EG95:amai},
evaluating {\em quantified Boolean formulas}~\cite{Stockmeyer77:tcs},
and {\em planning}~\cite{Niemela99:amai} for which 
encodings of the problem instances as logic
programs and test results are presented in
Sections~\ref{section:mm}--\ref{section:planning}, respectively. 
All benchmarks used in the experiments are available at
\url{http://www.tcs.hut.fi/Software/gnt/benchmarks/jnssy-tests-2003.tgz}.

\subsection{IMPLEMENTATION}
\label{section:implementation}

The implementation of \sys{GnT} \cite{GNT} is based on
\sys{smodels}~\cite{SMODELS,SNS02:aij}, a program that computes stable
models of normal logic programs. The basic idea behind \sys{GnT} is to
use two instances of the 
\sys{smodels} engine, one that generates the model candidates and one
that checks if they are minimal. To implement the idea it is enough to
extend the \sys{smodels} engine only slightly. Figure~\ref{fig:gnt}
shows the pseudo-code for \sys{GnT} modified from the original
$\smodels$ function presented in~\cite{SNS02:aij}.  The function
$\gnt(\GP,P,A)$ takes as input a normal (generator) program $\GP$, a
disjunctive program $P$ and a partial model (a set of literals) $A$
and performs a backtracking search for stable models of $\GP$.  It
returns a stable model $M$ of $\GP$ which agrees with $A$ and for
which $\isminimal(P,M)$ returns true if such a stable model exists and
otherwise it returns false.  It uses functions $expand(\GP,A)$,
$\extend(\GP,A)$, $\conflict(\GP,A)$, $\heuristic(\GP,A)$, and
$\isminimal(P,A)$. The first four are as in the original $\smodels$
procedure:
\begin{itemize}
\item
$\expand(\GP,A)$ returns a partial model which expands the given
partial model $A$ by literals satisfied by all (total) stable models
of $\GP$ agreeing with $A$ (obtained using a generalized well-founded
computation);

\item
$\extend(\GP,A)$ returns a partial model extending the partial model
$A$ by literals obtained by $\expand$ enhanced with lookahead
techniques.

\item
$\conflict(\GP,A)$ checks whether there is an immediate conflict,
i.e., if the partial model $A$ contains a complementary pair of
literals and

\item
$\heuristic(\GP,A)$ returns an atom undefined in $A$ to be used as the
next choice point in the backtracking search for stable models.
\end{itemize}
For further details on these functions see~\cite{SNS02:aij}.  The
function $\isminimal(P,A)$ performs the minimality test for a
disjunctive program $P$ and a partial model $A$ given in
Proposition~\ref{prop:early-test} using a call to \sys{smodels}, i.e.,
it views $A$ as a total model $A'$ where all atoms undefined in $A$
are taken to be false, builds the program $\test{P}{A'}$, calls
\sys{smodels} and returns false if $\test{P}{A'}$ has a stable model
and otherwise returns true.  To compute a stable model for a
disjunctive program $P$, the procedure $\gnt(\gen{P},P,\emptyset)$ is
called.  First $\gnt$ extends the given partial model and checks for
conflicts. If all atoms are covered by the extended partial model,
then a (total) model candidate has been found and it is checked for
minimality.  Otherwise the heuristic function selects a new undefined
atom $x$ and $\gnt$ searches recursively first for models where $x$ is
false.  If no such model is found, the partial model is expanded by
making $x$ true. If there is a conflict or the expanded model does not
pass an ``early'' minimality test, the procedure backtracks and
otherwise it continues the search recursively using the expanded
model. As the ``early'' minimality tests are computationally quite
expensive, some optimization has been employed so that
such tests are performed only when backtracking from 
a model candidate. For this there is a global
variable 'WasCovered' which is initially set to false and which is set
to true when a model candidate is found.  
However, it should be noticed that when backtracking from a model
candidate, the test could be repeated at each backtracking level until
it succeeds. 
The implementation of the $\gnt$ procedure shown in Figure
\ref{fig:gnt} consists of a few hundred lines of code~\cite{GNT} on
top of the \sys{smodels} system.

\begin{figure}
\begin{tabbing}
xx\=xx\=xx\=xx\=xx\=xx\=xx\=\kill
\textbf{function} $\gnt(\GP,P,A)$ \\
$A:=\extend(\GP,A)$ \\
\keyw{if} {$\conflict(\GP,A)$} \keyw{then}  \\
\> return false \\
\keyw{else if} {$A$ covers $\hb{\GP}$} \keyw{then}  \\
\> WasCovered := true \\
\> \keyw{if} {$\isminimal(P,A)$ } \keyw{then}  \\
\> \> return $A$ \\
\> \keyw{else}  \\
\> \> return false \\
\> \keyw{end if} \\
\keyw{else}\\
\> $x:=\heuristic(\GP,A)$\\
\> $A'$ := $\gnt(\GP,P,A\cup \{\naf x\})$ \\
\> \keyw{if} $A' \not= false$ \keyw{then} \\
\>\> return $A'$ \\
\> \keyw{else} \\
\>\> $A' := \expand(\GP,A\cup \{x\})$ \\
\>\> \keyw{if} $\conflict(\GP,A')$ \keyw{then} \\
\>\>\>  return false \\
\>\> \keyw{else if} WasCovered \keyw{then} \\
\>\>\> \keyw{if} not $\isminimal(P,A')$ \keyw{then} \\
\>\>\>\>  return false \\
\>\>\> \keyw{end if}\\
\>\> \keyw{end if} \\
\>\> WasCovered := false \\
\>\> return $\gnt\bigl(\GP,P,A'\bigr)$\\
\> \keyw{end if}\\
\keyw{end if}.
\end{tabbing}
\caption{\sys{GnT} Procedure}
\label{fig:gnt}
\end{figure}

In the sequel, we report several experiments which we carry out in
order to compare \sys{dlv} (version 2003-05-16) with \sys{GnT}
which is based on \sys{smodels} (version 2.27) and
uses \sys{lparse} (version 1.0.13) as an instantiator.  
We consider two versions
of our approach, \sys{GnT1} and \sys{GnT2}, which are similar except
that in \sys{GnT1} generating program $\genI{P}$ is used and in
\sys{GnT2}, $\gen{P} =\genI{P} \cup \support{P}$.
All of our tests are run under Linux 2.4.20 operating system on a
1.7 GHz AMD Athlon XP 2000+ computer with $1$ GB of memory. Execution
times are measured using the customary Unix \texttt{/usr/bin/time}
command.

\subsection{MINIMAL MODELS}
\label{section:mm}

Our first test problem is the $\Sigma^p_2$-complete problem of
deciding the existence of a minimal model of a set of clauses in which
some specified atoms are true \cite{EG95:amai}.
This problem is mapped to a stable model computation problem as
follows.  For a problem instance consisting of a set of clauses and
some specified atoms, a program $P$ is constructed where each clause
$a_1\lor\dotsb\lor a_n\lor\neg b_1\lor\dotsb\lor \neg b_m$ is
translated into a rule $a_1\lor\dotsb\lor a_n \IF b_1,\dotsc,b_m$ and
for each specified atom $c_i$, a rule
\begin{equation}
f\IF\naf f,\naf c_i\qquad 
\label{eq:true-atoms}
\end{equation}
is included. Now $P$ has a stable model if and only if there is a
minimal model of the clauses containing all specified atoms $c_i$. 

The test cases (random disjunctive 3-SAT programs) are based on random
3-SAT problems having a fixed clauses/atoms ratio $c$ and they are
constructed as follows.  Given a number of atoms $n$, a random 3-SAT
problem is generated, i.e.\ $c\times n$ clauses are generated each by
picking randomly three distinct atoms from the $n$ available and
selecting their polarity uniformly.  This is done using a program
\sys{makewff} developed by Bart Selman.  Then the clauses are
translated into rules as described above and for $i=1,\dotsc,\lfloor
2n/100\rfloor$ and for random atoms $c_i$, the extra rules
(\ref{eq:true-atoms}) are added.
The problem size is controlled by the number of atoms $n$ which is
increased by increments of $10$. For each $n$, we test $100$ random
3-SAT programs and measure the maximum, average, and minimum time it
takes to decide whether a stable model exists.

In the first set of tests we study the effect of different generating
programs on the performance of our approach, i.e., we compare
\sys{GnT1} and \sys{GnT2}, which are similar except
that in \sys{GnT1} generating program $\genI{P}$ is used and in
\sys{GnT2}, $\gen{P} =\genI{P} \cup \support{P}$.  We test at two
clauses/atoms ratios.  The first test is at 4.258 which is in the
phase transition region~\cite{CA96:aij} where roughly 50\% of the
generated 3-SAT clause sets are satisfiable.  The second test is at
clauses/atoms ratio 3.750 where practically all generated 3-SAT clause
sets are satisfiable.

\begin{figure*}
\begin{center}
\begin{tabular}{c}
\includegraphics{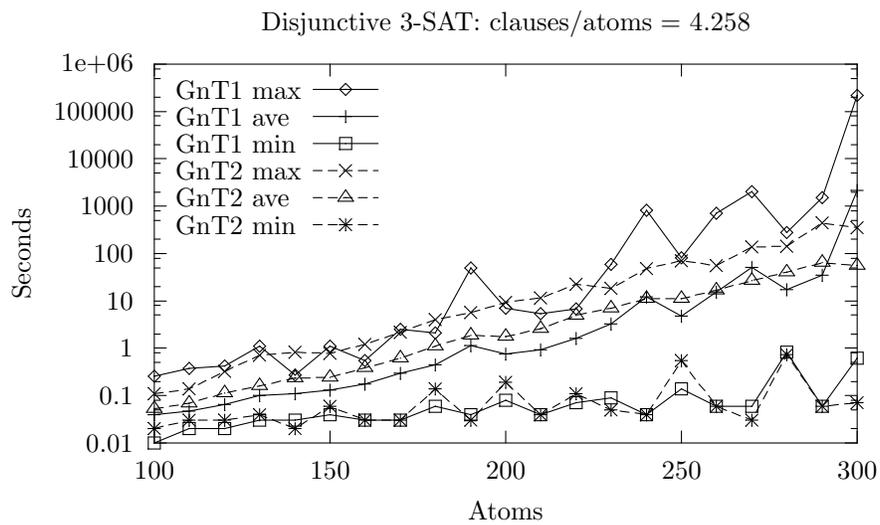} \\
\ \\
\ \\
\includegraphics{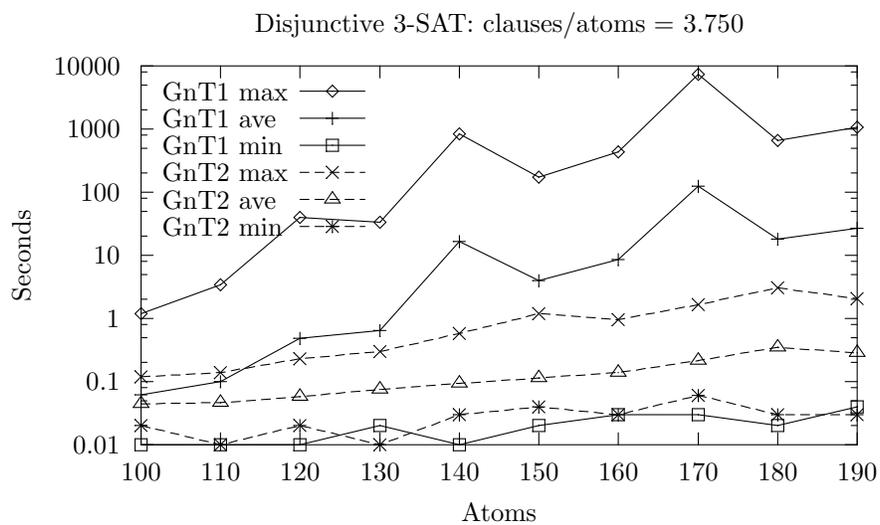}
\end{tabular}
\caption{Minimal Models: \sys{GnT1} vs.\ \sys{GnT2}}
\label{fig:expI}
\end{center}
\end{figure*}

\begin{figure*}
\begin{center}
\begin{tabular}{c}
\includegraphics{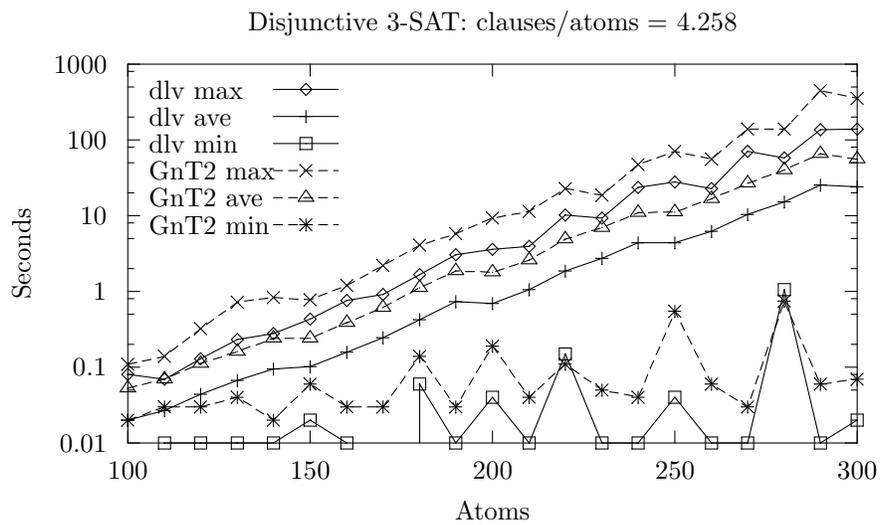} \\
\ \\
\ \\
\includegraphics{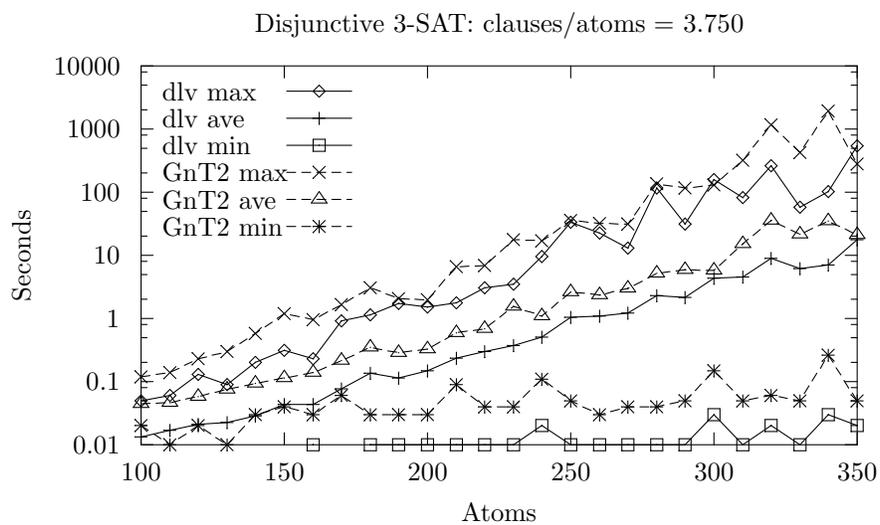}
\end{tabular}
\caption{Minimal Models: \sys{dlv} vs.\ \sys{GnT2}}
\label{fig:expII}
\end{center}
\end{figure*}

The test results are shown in Figure~\ref{fig:expI}. In the first test
set the key problem seems to be finding at least one model candidate.
The simpler generator (\sys{GnT1}) appears to perform relatively well except
for a few outliers, i.e.\ instances with significantly higher running
time than the average. The outliers occur when the generator program
$\genI{P}$ allows a high number of candidate models.  At clauses/atoms
ratio 3.750 the frequency of outliers for \sys{GnT1} increases and outliers
occur already in smaller problem sizes.  The more involved generating
program $\gen{P}$ behaves in a much more robust way and the average
running time of \sys{GnT2} is significantly lower than that of \sys{GnT1}.
Next we use the same two test sets for comparing \sys{GnT2} and
\sys{dlv}.  The results are shown in Figure~\ref{fig:expII}.  The
systems scale very similarly in both test sets but \sys{dlv} seems to
be roughly a constant factor faster than \sys{GnT2}. This is probably
due to the overhead caused by the more complicated generating program
in \sys{GnT2} and by the two level architecture of \sys{GnT2} where
two instances of \sys{smodels} are cooperating.

\subsection{QUANTIFIED BOOLEAN FORMULAS}

We continue the comparison of \sys{GnT2} and \sys{dlv} using instances
of quantified Boolean formulas (QBFs) and develop a new way to encode
such formulas as disjunctive logic programs.  In our experiments, we
consider a specific subclass of QBFs, namely $2,\exists$-QBFs. Such
formulas are of the form $\exists X\forall Y\phi$ where $X$ and $Y$
are sets of existentially and universally quantified propositional
variables, respectively, and $\phi$ is a Boolean formula based on
$X\union Y$.
Deciding the validity of such a formula forms a $\Sigma^p_2$-complete
decision problem \cite{Stockmeyer77:tcs} even if $\phi$ is assumed to
be a Boolean formula in 3DNF \cite{SM73:stoc}.  Recall that checking
the existence of a stable model for a disjunctive logic program is of
equal computational complexity \cite{EG95:amai}, which implies the
existence of polynomial time transformations between the decision
problems mentioned above.
In fact, Eiter and Gottlob \cite{EG95:amai} show how a QBF of the form
$\exists X\forall Y\phi$ with $\phi$ in 3DNF can be translated into
a disjunctive logic program $P$ such that $\exists X\forall Y\phi$ is
valid if and only if $P$ has a stable model. This translation is used
by Leone et al.~\cite{LPFEGPS03:submitted} to compare \sys{dlv} and
\sys{GnT2}.

However, we present an alternative transformation in order to obtain a
better performance for the two systems under comparison. Our
transformation is based on the following ideas. The first observation
is that we can rewrite $\exists X\forall Y\phi$ with $\phi$ in DNF as
$\exists X\neg\exists Y\neg\phi$ where $\neg\phi$ can be understood as
a Boolean formula in CNF or as a {\em set of clauses} $S$ so that each
clause $c\in S$ can represented as a disjunction of the form
\begin{equation}
\label{eq:clause}
X_1\lor\neg X_2\lor Y_1\lor\neg Y_2
\end{equation}
where $X_1$ and $\neg X_2$ are the sets of positive and negative
literals, respectively, which appear in $c$ and involve variables from
$X$ while $Y_1$ and $\neg Y_2$ are similarly related to $Y$.
It follows that $\exists X\forall Y\phi$ is valid if and only if we
can find an interpretation\footnote{The values assigned by $I$ to the
variables in $Y$ are not important, but make $I$ a proper
interpretation over $X\union Y$.} $\func{I}{X\union
Y}{\set{\true,\false}}$ such that $\neg\phi_{I(X)}$ is unsatisfiable
where $\neg\phi_{I(X)}$ denotes the set of clauses $Y_1\lor\neg Y_2$
for which (\ref{eq:clause}) belongs to $S$ and $I\not\models
X_1\lor\neg X_2$.
The second idea behind our transformation is to choose the truth value
of the condition $X_1\lor\neg X_2$ for each clause (\ref{eq:clause})
in $S$ rather than the truth values of the variables in $X\union Y$.
This line of thinking leads to the following translation of $S$.

\begin{definition}
\label{def:QBF-transformation}
A clause $c$ of the form (\ref{eq:clause}) where $X_1,X_2\subseteq X$
and $Y_1,Y_2\subseteq Y$ is translated into following sets of rules:
\begin{center}
$\begin{array}{r@{\ }c@{\ }l}
\tr{V}{c} & = & \set{c\IF\naf\hat{c}\END
                     \hat{c}\IF\naf c}, \\
\tr{E}{c} & = & \sel{f\IF x,\naf\hat{c},\naf f}{x\in X_1}\union
                \sel{x\IF\naf\hat{c}}{x\in X_2}\ \union \\
          &   & \set{f\IF X_2,\naf X_1,\naf c,\naf f},\text{ and} \\
\tr{U}{c} & = & \sel{y\IF u}{y\in Y_1\union Y_2}\union
                \set{Y_1\union\set{u}\IF Y_2,\naf \hat{c}}
\end{array}$
\end{center}
where $c$ and $\hat{c}$ are new atoms associated with the clause $c$,
and $f$ and $u$ are new atoms. A set of clauses $S$ is translated into
\begin{center}
$\Union_{c\in S}(\tr{V}{c}\union\tr{E}{c}\union\tr{U}{c})
 \union\set{u\IF\naf u}$.
\end{center}
\end{definition}

We use Boolean variables from $X\union Y$ as propositional atoms in
the translation.
Intuitively, the rules of $\tr{V}{c}$ choose whether a clause $c$ is
{\em active}, i.e.\ $X_1\lor\neg X_2$ evaluates to false so that the
satisfaction of the clause $c$ depends on the values assigned to
$Y_1\union Y_2$.  The rules in $\tr{E}{c}$ try to {\em explain} the
preceding choice by checking that the values of the variables in $X$
can be assigned accordingly. Finally, the rules in $\tr{U}{c}$
implement the test for unsatisfiability together with the rule
$u\IF\naf u$.
Basically, the same unsatisfiability check is used in the translation
proposed by Eiter and Gottlob. However, the transformation given in
Definition \ref{def:QBF-transformation} is more economical as it uses
far less new atoms and disjunctive rules. In particular, note that
variables from $X\union Y$ not appearing in the clauses do not
contribute any rules to the translation.

Next we address the correctness of our transformation and consider a
$2,\exists$-QBF $\exists X\forall Y\phi$ where $\phi$ is in DNF and
the disjunctive logic program $P$ obtained by translating $\neg\phi$
(a set of clauses $S$) according to Definition
\ref{def:QBF-transformation}.

\begin{lemma}
\label{lemma:reduct}
Let $M\subseteq\hb{P}$ be a total propositional interpretation for the
translation $P$ such that for every clause $c\in S$ of the form
(\ref{eq:clause}), (i) $c\in M$ $\iff$ $\hat{c}\not\in M$ and (ii)
$c\in M$ $\iff$ $X_1\isect M=\emptyset$ and $X_2\subseteq M$.

Then the programs $\tr{V}{c}$,
$\tr{E}{c}$, and $\tr{U}{c}$ associated with a clause $c\in S$ of the
form (\ref{eq:clause}) satisfy the following.
\begin{enumerate}
\item[(R1)]
The fact $c\IF$ belongs to $\GLred{\tr{V}{c}}{M}$ $\iff$ $c\in M$.

\item[(R2)]
The fact $\hat{c}\IF$ belongs to $\GLred{\tr{V}{c}}{M}$ $\iff$
$\hat{c}\in M$.

\item[(R3)]
For $x\in X_1$, the rule $f\IF x$ belongs to $\GLred{\tr{E}{c}}{M}$ $\iff$ \\
$x\not\in M$ and $f\not\in M$.

\item[(R4)]
For $x\in X_2$, the fact $x\IF$ belongs to
$\GLred{\tr{E}{c}}{M}$ $\iff$ $x\in M$.

\item[(R5)]
The rule $f\IF X_2$ belongs to
$\GLred{\tr{E}{c}}{M}$ $\iff$ $X_2\not\subseteq M$ and $f\not\in M$.

\item[(R6)]
For $y\in Y_1\union Y_2$, the rule $y\IF u$ belongs to
$\GLred{\tr{U}{c}}{M}$ unconditionally.

\item[(R7)]
The rule $Y_1\union\set{u}\IF Y_2$ belongs to $\GLred{\tr{U}{c}}{M}$
$\iff$ $c\in M$.
\end{enumerate}
\end{lemma}

\begin{theorem}
The quantified Boolean formula $\exists X\forall Y\phi$ is valid if
and only if the translation $P$ has a stable model.
\end{theorem}

\begin{proof}
We may safely assume that all variables in $X\union Y$ actually appear
in $\phi$, since redundant variables can be dropped without affecting
the validity of the formula nor the structure of its translation.

($\implies$)
Suppose that $\exists X\forall Y\phi$ is valid. Then there is an
interpretation $\func{I}{X\union Y}{\set{\true,\false}}$ such that
$I\models\forall Y\phi$.
Then define $X_I=\sel{x\in X}{I(x)=\true}$.
Without loss of generality we may assume that $X_I$ is minimal, i.e.\
there is no interpretation $J$ such that $J\models\forall Y\phi$ and
$X_J\subset X_I$. Then define a total propositional interpretation
\begin{multline}
\label{eq:stable-model-structure}
M=X_I\union Y\union\set{u}\ \union \\
\sel{c}{c=X_1\lor\neg X_2\lor Y_1\lor\neg Y_2\in\neg\phi
        \text{ and }I\not\models X_1\lor\neg X_2}\ \union \\
\sel{\hat{c}}{c=X_1\lor\neg X_2\lor Y_1\lor\neg Y_2\in\neg\phi
              \text{ and }I\models X_1\lor\neg X_2}.
\end{multline}
It is verified next that $M$ is a stable model of $P$. The definition
of $M$ implies that $M$ satisfies the requirements of Lemma
\ref{lemma:reduct}.  Then (R1)--(R7) effectively describe the
structure of $\GLred{P}{M}$ and it is easy to verify that $M$ is a
model of $\GLred{P}{M}$ on the basis of these relationships, as
$Y\union\set{u}\subseteq M$ and $f\not\in M$ by definition.

Next we assume that $N\subseteq M$ is a model of $\GLred{P}{M}$
and show that $M\subseteq N$.
(i)
If $c\in M$ for some clause $c$ of the form (\ref{eq:clause}),
then $c\in N$, as $c\IF$ belongs to $\GLred{P}{M}$ by (R1).
(ii)
Similarly, $\hat{c}\in M$ implies $\hat{c}\in N$ by (R2).
(iii)
We have $u\in N$ because otherwise $N$ would form a model of
$\neg\phi_{I(X)}$ by satisfying the rules $Y_1\union\set{u}\IF Y_2$
included in $\GLred{P}{M}$ by (R7).
(iv)
Moreover, $Y\subseteq N$ holds, as $u\in N$ and $N$ satisfies all the
rules $y\IF u$ belonging to $\GLred{P}{M}$ by (R6).
(v)
Let us define an interpretation $\func{J}{X\union
Y}{\set{\true,\false}}$ such that for $x'\in X$, $J(x')=\true$ $\iff$
$x'\in N$, and for $y\in Y$, $J(y)=I(y)$. Using (R4), we can establish
for any (\ref{eq:clause}) that $I\not\models X_1\lor\neg X_2$ implies
$J\not\models X_1\lor\neg X_2$.  Thus
$\neg\phi_{I(X)}\subseteq\neg\phi_{J(X)}$ where $\neg\phi_{I(X)}$ is
known to be unsatisfiable.  The same follows for $\neg\phi_{J(X)}$ so
that $J$ qualifies as an assignment for which $J\models\forall Y\phi$
holds. But then the minimality of $I$ implies $J=I$, $X_J=X_I$, and
$X_I\subseteq N$.
To conclude the preceding analysis,
$M\subseteq N$ and $M$ is a stable model of $P$.

($\impliedby$)
Suppose that $P$ has a stable model $M$.
Then define an interpretation $\func{I}{X\union
Y}{\set{\true,\false}}$ by setting $I(z)=\true$ $\iff$ $z\in M$ for
any $z\in X\union Y$. Let us then establish that $M$ and $I$
satisfy (\ref{eq:stable-model-structure}).
(i)
The definition of $I$ implies that $X_I=M\isect X$.
(ii)
Now $u\in M$, because $P$ contains $u\IF\naf u$ and
$M$ is a stable model of $P$.
(iii)
For the same reason, $f\not\in M$, because all the rules having
$f$ as the head have $\naf f$ among the negative body literals.
(iv)
Since $u\in M$ and $\GLred{P}{M}$ contains the rule $y\IF u$ for every
$y\in Y$, we obtain $Y\subseteq M$.
(v)
For any clause $c$ of the form (\ref{eq:clause}), the structure of
$\tr{V}{c}\subseteq P$ implies that $c\in M$ $\iff$ $c\IF$ belongs to
$\GLred{\tr{V}{c}}{M}\subseteq\GLred{P}{M}$ $\iff$ $\hat{c}\not\in M$.
Using this property, we can establish that $c\in M$ $\iff$
$I\not\models X_1\lor\neg X_2$ holds for the interpretation $I$
defined above.
(vi)
Thus $\hat{c}\in M$ $\iff$ $I\models X_1\lor\neg X_2$ is implied by
the fact that $c\in M$ $\iff$ $\hat{c}\not\in M$, as shown above in (v).

It remains to show that $\neg\phi_{I(X)}$ is unsatisfiable. So let us
assume the contrary, i.e.\ there is a model $Y'\subseteq Y$ for
$\neg\phi_{I(X)}$. Note that $M$ meets the requirements of Lemma
\ref{lemma:reduct} by (v) and (vi) above, as the definition of $I$
implies $I\not\models X_1\lor\neg X_2$ $\iff$ $X_1\isect M=\emptyset$
and $X_2\subseteq M$.
The relationships (R1)--(R7) imply that $N=(M-(Y\union\set{u}))\union
Y'$ is a model of $\GLred{P}{M}$, too.  Since $u\not\in N$, we have
$N\subset M$ indicating that $M$ is not stable, a contradiction. Thus
$\neg\phi_{I(x)}$ is unsatisfiable which implies $I\models\forall
Y\phi$ and the validity of $\exists X\forall Y\phi$.
\end{proof}

\begin{figure*}
\begin{center}
\begin{tabular}{c}
\includegraphics{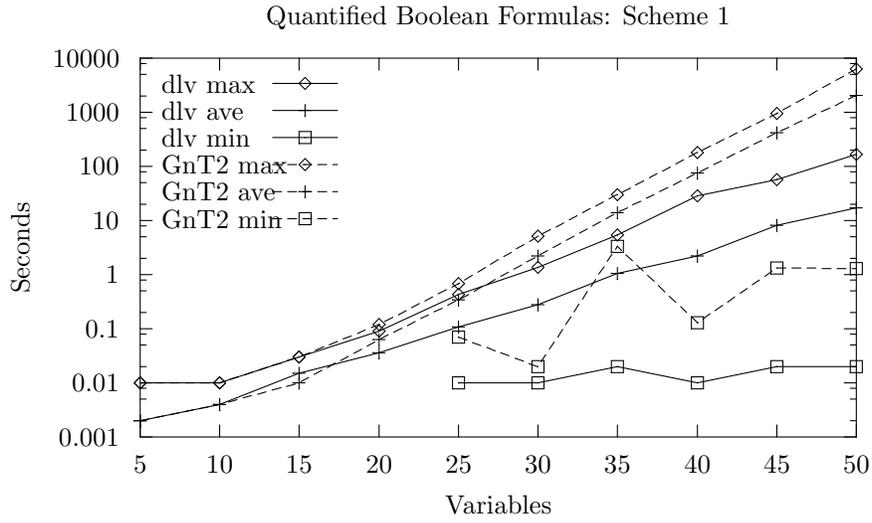} \\
\ \\
\ \\
\includegraphics{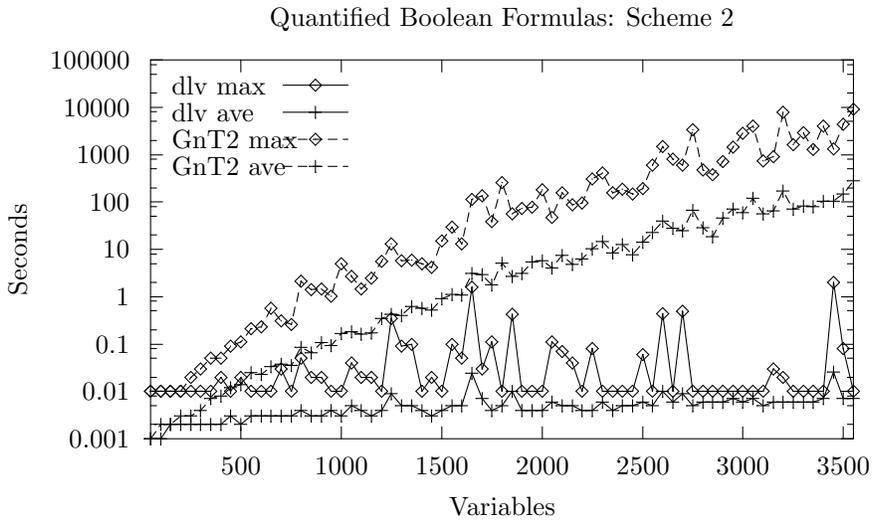}
\end{tabular}
\end{center}
\caption{Quantified Boolean Formulas: \sys{dlv} vs.\ \sys{GnT2}}
\label{fig:expIII}
\end{figure*}

Using an implementation of the translation given in Definition
\ref{def:QBF-transformation} we are able to transform $2,\exists$-QBFs
into disjunctive programs. The remaining question is how to generate
$2,\exists$-QBF instances. We use two different schemes based on
random instances \cite{CGS98:aaai,GW99:aaai}.
In the first scheme, the sets of variables $X$ and $Y$ satisfy
$|X|=|Y|$. Each random instance is based on $v=|X|+|Y|$ variables and
a Boolean formula $\phi$ which is a disjunction of $d=2\times v$
conjunctions of $5$ random literals out of which at least two literals
involve a variable from $Y$, as suggested by Gent and Walsh
\cite{GW99:aaai}. This scheme is slightly different from
$\mathrm{2QBF}_{GW}$ in \cite{LPFEGPS03:submitted} based on $3$
literal conjunctions just to obtain a more challenging
benchmark.
The constant factor $2$ in the equation relating $d$ and $v$ has been
determined as a phase transition point for the \sys{dlv} system by
keeping $v=50$ fixed and varying the number of disjunctions in
$\phi$. In the actual experiment, the number of $v$ variables is
varied from $5$ to $50$ by increments of $5$.
We generate $100$ instances of $2,\exists$-QBFs for each value of $v$
and translate them into corresponding disjunctive logic programs.  The
running times for \sys{dlv} and $\sys{GnT2}$ are depicted in the upper
graph of Figure \ref{fig:expIII}. The systems scale very similarly,
but \sys{dlv} is on the average from one to two decades faster than
\sys{GnT2}.

In the second experiment with $2,\exists$-QBFs, we use a different
scheme for the number of disjunctions $d=\lfloor\sqrt{v/2}\rfloor$ as
well as the number of literals which is $3$ in each conjunction.  The
resulting instances are much easier to solve, because $d$ remains
relatively low (e.g.\ $d\approx 41$ for $v=3500$) and many variables
do not appear in $\phi$ at all. We let $v$ vary from $50$ to $3550$ by
increments of $50$ and generate $100$ instances of $2,\exists$-QBFs
for each value of $v$. The resulting running times are shown in the
lower graph of Figure \ref{fig:expIII}.  The shapes of the curves are
basically the same, but the performance of \sys{GnT2} degrades faster
than that of \sys{dlv}.
However, the benefits of the translation given in Definition
\ref{def:QBF-transformation} are clear, as \sys{GnT2} is able to solve
much larger instances than reported in \cite{LPFEGPS03:submitted}
where $40$ variables turn out to be too much for \sys{GnT2}.  As far
as we understand, this is due to the sizes of search spaces associated
with the translated instances of $2,\exists$-QBFs. For the translation
given in Definition \ref{def:QBF-transformation}, the size of the
search space examined by \sys{GnT2} is of order $2^{\sqrt{v/2}}$
whereas it is of order $2^v$ if the translation proposed by Eiter and
Gottlob \cite{EG95:amai} is used.

\subsection{PLANNING}
\label{section:planning}

In order to get an idea of the overhead of \sys{GnT2} when compared to
\sys{smodels}, we study three blocks world planning problems encoded
as normal programs~\cite{Niemela99:amai}:
\begin{itemize}
\item
{large.c} is a 15 blocks problem requiring a 8 step plan using the
encoding given in~\cite{Niemela99:amai} allowing parallel execution of
operators,

\item
{large.d} is a 17 blocks problem with a 9 step plan and

\item
{large.e} is a 19 blocks problem with a 10 step plan.
\end{itemize}
Table~\ref{table:bw-results} contains two entries for each problem:
one reporting the time needed to find a valid plan with the
``optimal'' number of steps given as input and one reporting the time
needed to show optimality, i.e., that no plan (no stable model) exists
when the number of situations is decreased by one.  The times reported
for each test case are the execution times of \sys{smodels} and
\sys{GnT2} given a ground normal program (generated by \sys{lparse})
as input. The results show that there is some overhead in the current
implementation of \sys{GnT2} even for normal programs and \sys{GnT2}
handles these examples 2-4 times slower than \sys{smodels}.

\begin{table}
\centering
\caption{Planning: \sys{smodels} vs.\ \sys{GnT2}}
\label{table:bw-results}

\begin{tabular}{lrrrr} \hline
Problem & Number of  & Number of    & Time (s) & Time (s)\\
        & steps & ground rules    & \sys{smodels} & \sys{GnT2} \\ \hline
large.c & 8          &  81681       & 4.5  & 10.3 \\
        & 7          & 72527     & 0.6     & 2.1 \\ \hline
large.d & 9          &  127999      & 10.1 & 21.2 \\
        & 8          &  115109       & 1.4 & 5.2\\ \hline
large.e & 10         &  191621     & 18.2  & 35.0\\
        & 9         &  174099  & 2.2       & 8.7\\ \hline
\end{tabular}
\end{table}


\section{CONCLUSIONS}
\label{section:conclusions}

The paper presents an approach to implementing partial and disjunctive
stable models using an implementation of stable models for
disjunction-free programs as the core inference engine.  The approach
is based on unfolding partiality and disjunctions from a logic program
in two separate steps.  In the first step partial stable models of
disjunctive programs are captured by total stable models using a
simple linear program transformation.  Thus, reasoning tasks
concerning partial models can be solved using an implementation of
total models such as the \sys{dlv} system.  This also sheds new light
on the relationship between partial and total stable models by
establishing a close correspondence.  In the second step a generate
and test approach is developed for computing total stable models of
disjunctive programs using a core engine capable of computing stable
models of normal programs. We have developed an implementation of the
approach using \sys{smodels} as the core engine. The extension is
fairly simple consisting of a few hundred lines of code.  The approach
turns out to be competitive even against a state-of-the-art system for
disjunctive programs.  The efficiency of the approach comes partly
from the fact that normal programs can capture essential properties of
disjunctive stable models that help with decreasing the computational
complexity of the generate and test phases in the approach. However, a
major part of the success can be accounted for by the efficiency of
the core engine. This suggests that more efforts should be spent in
developing efficient core engines.

\subsubsection*{ACKNOWLEDGEMENTS} 
 
The work of the first, the second and the fourth author has been
funded by the Academy of Finland (projects \#43963 and \#53695), and
that of the fourth author also by the Helsinki Graduate School in
Computer Science and Engineering.  We thank Tommi Syrjänen for
implementing the front-end \sys{lparse} for the \sys{smodels}
\cite{SMODELS} system and for incorporating the relevant translations
into the implementation.  We also thank Emilia Oikarinen who
implemented the transformation from QBFs to disjunctive programs.


\renewcommand{\refname}{REFERENCES}


\end{document}